\documentclass[10pt,twocolumn,letterpaper]{article}
\usepackage{3dv}
\usepackage{times}
\usepackage{epsfig}
\usepackage{graphicx}
\usepackage{amsmath}
\usepackage{amssymb}
\usepackage{textcomp}
\usepackage{gensymb}
\usepackage{caption}
\usepackage{multirow}
\usepackage{tikz}
\usepackage{pgfplots}
\usepackage{xspace}
\pgfplotsset{compat=1.13}

\usepackage[pagebackref=true,breaklinks=true,colorlinks,bookmarks=false]{hyperref}

\threedvfinalcopy 
\def\threedvPaperID{124} 

\ifthreedvfinal\fi

\begin{document}

\title{A Data-driven Prior on Facet Orientation for Semantic Mesh Labeling}

\author{Andrea Romanoni\\
Politecnico di Milano, Italy\\
{\tt\small andrea.romanoni@polimi.it}
\and
Matteo Matteucci\\
Politecnico di Milano, Italy\\
{\tt\small matteo.matteucci@polimi.it}
}

\maketitle

\begin{abstract}
Mesh labeling is the key problem of classifying the facets of a 3D mesh with a label among a set of possible ones. State-of-the-art methods model  mesh labeling as a Markov Random Field over the facets. These algorithms map image segmentations to the mesh by minimizing an energy function that comprises a data term, a smoothness terms, and class-specific priors. The latter favor a labeling with respect to another depending on the orientation of the facet normals. In this paper we propose a novel energy term that acts as a prior, but does not require any prior knowledge about the scene nor scene-specific relationship among classes. It bootstraps from a coarse mapping of the 2D segmentations on the mesh, and it favors the facets to be labeled according to the statistics of the mesh normals in their neighborhood. We tested our approach against five different datasets and, even if we do not inject prior knowledge, our method adapts to the data and overcomes the state-of-the-art.
\end{abstract}

\section{Introduction}
In the last decades we witnessed a significant evolution in 3D modeling from images: moving from the early point-based approaches, researchers have proposed a wide variety of solutions that output a dense 3D mesh model of the environment.
The most recent trend in computer vision, as well as in robotics and photogrammetry, is to add semantic labels to the 3D reconstruction, in order to estimate a more informative 3D model for robot navigation and localization or visualization and inspection.

\begin{figure}
\centering
\includegraphics[width=0.85\columnwidth]{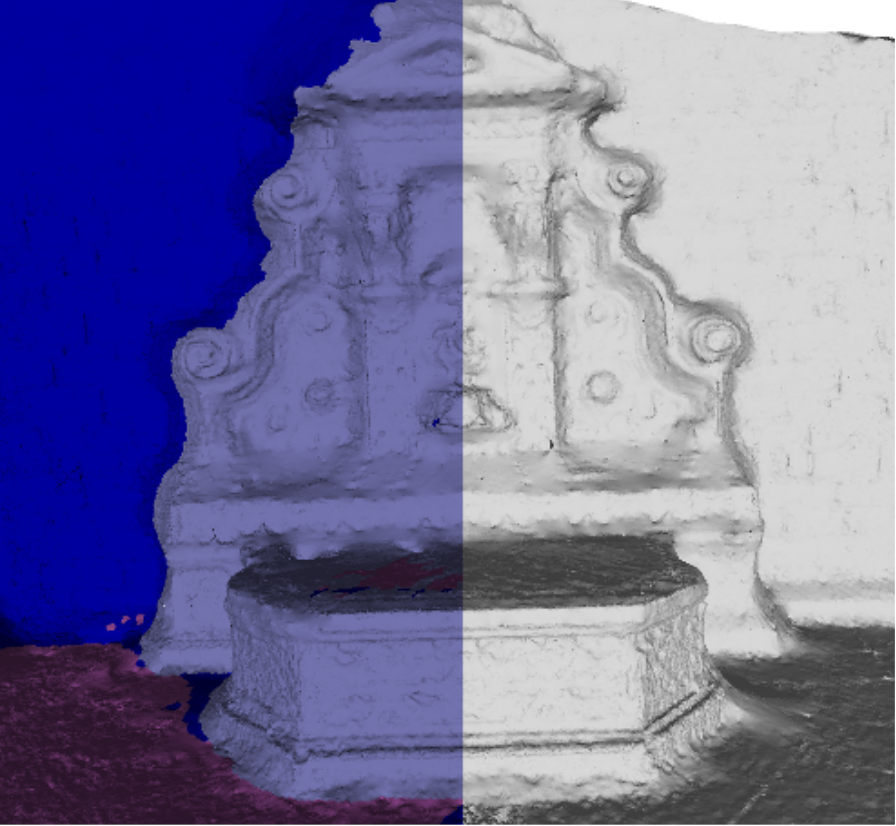}
\caption{Example of semantic mesh labeling using the method proposed in this paper on the fountain-P11 dataset}
\label{fig:segex}
\end{figure}

On one side volumetric semantic 3D reconstruction~\cite{hane2013joint,kundu2014joint,savinov2015discrete} couples the 3D reconstruction with semantic labeling; this improves the quality of the 3D model recovered  from the images, but the complex minimization techniques involved and the spatial complexity of the volumetric  representation affect negatively  the efficiency and the scalability of this approach.
Recent semantic Multi-View Stereo (MVS) mesh refinement address the problem by estimating a low resolution mesh and refining it exploiting photometric and semantic information~\cite{blaha2017semantically,romanoni2017multi}. As the shape and resolution of the mesh change while the algorithm refines the mesh, the  labels need to be updated accordingly. Therefore a time-crucial step is the semantic mesh labeling.
Moreover, when a big and accurate 3D model, recovered from image-based algorithms, such as~\cite{vu_et_al_2012}, is already available, recomputing the entire model through expensive volumetric semantic 3D reconstruction algorithms is inconvenient or, in some cases, impossible due to memory restrictions.
In such cases mesh labeling represents the unique feasible possibility to recover a semantic reconstruction.

Semantic mesh labeling aims at classifying each facet of a 3D mesh model by exploiting available 2D images segmentation, 3D cues, or geometric priors, as in Figure \ref{fig:segex}.
Successful approaches leverage 2D image segmentations, and model the labeling problem as energy minimization  over a Markov Random Field (MRF).
In the MRF, the nodes represent facets whereas the potentials gather semantic information from image segmentations, enforce smoothing, and represent class-specific priors.
For instance, Blaha~\etal~\cite{blaha2017semantically} proposes a unary prior which, for each class, defines a preferred orientation with respect to the ground vector. 

In this paper we propose a novel MRF formulation that does not require any additional knowledge of the environment except the 3D model and the image segmentations. 
We propose to use a novel unary term in conjunction with the usual data term a smoothing term and the discontinuity term proposed in~\cite{romanoni2017multi}. 
We exploit a coarse labeling of the 3D mesh to collect locally the distribution of the normals for each class, it biases the facet classification according to them; instead of defining manually the preferred orientation for each class, as in~\cite{blaha2017semantically} or estimating it globally as in~\cite{romanoni2017multi}.

\begin{figure}[tp]
\centering
\begin{tabular}{cc}
\includegraphics[width=0.2\textwidth]{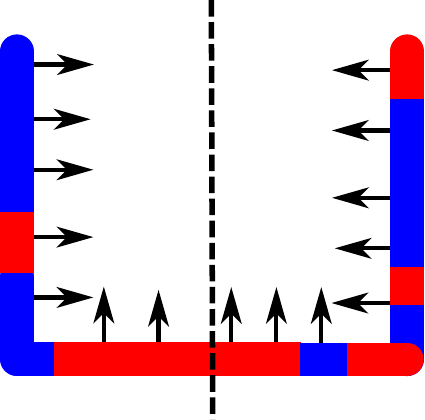}&
\includegraphics[width=0.2\textwidth]{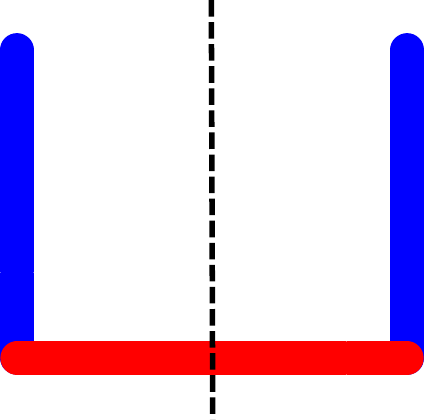}\\
(a)&
(b)\\
\end{tabular}
\caption{Toy example that shows the idea behind the term  $E_{\textrm{norm}}$}
\label{fig:Toy}
\end{figure}

\section{Related Works}
\label{sec:related}
Recovering a semantically annotated 3D model has been addressed by joining the labeling with the 3D reconstruction, or by decoupling the two tasks.

Joint semantic and 3D reconstruction has first been  proposed by H\"{a}ne \etal~\cite{hane2013joint} and it stems from the idea that scene semantics enables the definition of class-priors to improve the quality and the robustness of a 3D reconstruction algorithm. 
In their work, the authors define a convex minimization problem in the continuum 3D space with a unary and a pairwise potential, then they relax it to work in the discrete voxel space.
The per-voxel unary potential is estimated from 2D image segmentations and the visibility rays from each camera (data term).
The pairwise term is the combination of smoothness and class-specific priors, the latter is learned from manually annotated data, and defines the probability of class changes depending on the normal transition.
This work has been extended in~\cite{savinov2015discrete,savinov2016semantic} to embed higher order ray potentials  into the optimization of the graphical model.
For a more in-depth discussion about semantic volumetric reconstruction we refer the reader to~\cite{hane2016overview}.
Conversely, Kundu \etal~\cite{kundu2014joint} propose to bootstrap from Structure from Motion (SfM) points and they define a Conditional Markov Random Field on a 3D voxel grid, where they estimate the unary potentials from image segmentations, and they add handcrafted class-specific priors to enforce reconstruction consistency. 
The method is effective for large scale environment, but the 3D labeled model has a low resolution due to the low number of SfM points.
 
Even if joint segmentation and reconstruction algorithms are effective, they strongly rely on volumetric reconstruction and the optimization techniques are usually expensive so that they require a difficult trade off between accuracy scalability and computational speed.
More efficient representations have been proposed, \eg, octrees as in~\cite{sengupta2015semantic}, voxel-hashing~\cite{vineet2015incremental}, submaps~\cite{cherabier2016multi}, or multi-grids
\cite{blaha2016large}; however the scalability of multi-label optimization of these methods is still an open issue.

Different approaches decouple the reconstruction and the mesh labeling, so that the complexity required by the 3D reconstruction module is reduced, and the scene segmentation is delegated to a semantic labeling algorithm.
Sengupta \etal~\cite{sengupta2013urban} build a 3D volumetric reconstruction through the standard Truncated Signed Distance Function (TSDF), then they label the images and fuse them in the 3D model.
Instead, Valentin \etal~\cite{valentin2013mesh},  design a CRF whose unary potentials correspond directly to 3D voxels and are trained by taking into account images color, and geometric features of the facets. 
The authors improve the labeling with respect to Sengupta \etal~\cite{sengupta2013urban}, however the algorithm is computationally expensive and, in our paper, we show that the priors learning phase is not needed. 

Herman \etal~\cite{hermans2014dense} and McCormac \etal~\cite{mccormac2017semanticfusion} fuse incrementally the 2D images segmentation into the reconstruction. 
These methods, as well as~\cite{sengupta2013urban}, benefit from the vast literature about 2D segmentation algorithms for 2D image segmentation~\cite{chen2016deeplab,visin2016reseg,garcia2017review}. 
However, the output of~\cite{hermans2014dense} and~\cite{mccormac2017semanticfusion} are not continuous triangular meshes, but, respectively, a point cloud and a surfel model.
Tung \etal~\cite{tung2017mf3d} proposed an interesting method that estimates the unary potential of a CRF without any classifier, but by comparing the regions of the image computed as in~\cite{tung2014collageparsing}, with a database of annotated examples. However they do not use any semantic prior to add robustness of the final outcome. 

Kalogerakis \etal \cite{kalogerakis2010learning} segment meshes into semantics categories by learning  different 3D features of the mesh such as shape curvature.
Verdie \etal~\cite{verdie2015lod} take as input an existing  3D mesh of a city reconstructed from aerial images. 
They subdivide it into superfacets, which are subareas of the 3D mesh, and they collect geometric attributes, \ie, elevation, planarity and horizontality. 
Then they model the problem with a Markov Random Field with unary class-likelihood and pairwise class-transition terms defined manually.
Even if the results are remarkable, the handcrafted priors do not generalize the method in non urban areas.
For this reason Rouhani \etal~\cite{rouhani2017semantic} and Martinovic \etal~\cite{martinovic20153d} extended~\cite{verdie2015lod} with a classifier trained on both geometrical features and color information from the mesh texture. 
The method improves the results of~\cite{verdie2015lod} and it avoids using handcrafted priors; however, due to the features taken into account, its scope is still limited to city models from aerial images.
In~\cite{gadde2017efficient} the authors propose a series of classifiers that labels a point cloud but even in this case the algorithm
is domain-specific (restricted to fa\c{c}ades) and its application is restricted to building facades.  The main drawback of such methods that rely on just 3D cues to segment the mesh, is that they are not able to exploit the available deep learning methods such as~\cite{chen2016deeplab,visin2016reseg,garcia2017review}, that in the last years have shown impressive performances in image segmentation.
Riemenschneider \etal~\cite{riemenschneider2014learning} propose to classify a mesh according to 2D image segmentation, however their method requires a learning stage to estimate which surface orientation has to be labeled from which viewing direction by relying on 3D geometric features.

Blaha \etal ~\cite{blaha2017semantically} and Romanoni \etal~\cite{romanoni2017multi} propose two different mesh labeling methods, which are the closest to our proposal, and which rely on 2D image segmentation. 
Both~\cite{blaha2017semantically} and~\cite{romanoni2017multi} are two MVS algorithms that extend the photometric refinement of~\cite{vu_et_al_2012}.
They bootstrap from a 3D mesh and 2D image segmentations and they jointly refine the former through both semantics and photometric data. Mesh
labeling plays a substantial role here; since the model changes as new refinement iterations are applied, the mesh needs to be relabeled at each iteration. They define a Markov Random Field (MRF) over the mesh facets with unary potentials induced by the 2D image segmentations and pairwise smoothing terms for adjacent facets. 
Blaha \etal~\cite{blaha2017semantically} add a unary geometric term that encodes handcrafted class-specific priors; Romanoni \etal~\cite{romanoni2017multi} estimate per-class normal distributions from a coarse mesh labeling and they add a unary term that, if the facets labeled as c show a strong preference for a direction, other facets with similar direction are biased to belong to the same class c. The relevant aspect of these two methods and the method proposed in this paper is that they do not need to learn unary potentials over class labels.

\section{Proposed Method}
\label{sec:proposed}
The proposed algorithm models the labeling as an energy minimization problem over a Markov Random Field. 
The inputs are: the 3D triangular mesh model of the scene under analysis, and, for each image that captures the scene, the corresponding camera calibration and the output of a semantic classifier in the form of a likelihood distribution for each class $c$.

Given a label assignment $\mathbf{l}$, which associates  a label $l_f$ to each facet $f\in\mathcal{F}$, we define the following energy:
\begin{multline}
\label{eq:fullene}
E(\mathbf{l}) =  \sum_{f\in\mathcal{F}} E_{\textrm{data}}(l_f) + \mu_1 \sum_{f\in\mathcal{F}}E_{\textrm{norm}}(l_f) + \\ \mu_2  \sum_{f,h\in\mathcal{F}} E_{\textrm{disc}}(l_f,l_h) + \mu_3 \sum_{f,h\in\mathcal{F}} E_{\textrm{smooth}}(l_f) 
\end{multline}
where $E_{\textrm{data}}$ and $E_{\textrm{smooth}}$ are respectively the data unary term proposed in~\cite{blaha2017semantically} and a smoothness term adapted from~\cite{blaha2017semantically}.
The pairwise term $E_{\textrm{disc}}$, we named discontinuity prior,  was proposed in~\cite{romanoni2017multi} and biases the class changes between adjacent facets where significant variations between normal orientations occur.

The main contribution of this paper is the unary term $E_{\textrm{norm}}$ that replaces the priors defined manually as in~\cite{blaha2017semantically} or by means of learning algorithms as in~\cite{hane2013joint} or~\cite{rouhani2017semantic}.
Assuming a coarse labeling of the mesh is available, through the term $E_{\textrm{norm}}$  we label each facet by comparing its normal direction with the normal directions of the facets in its neighborhood. 
For instance, in Figure \ref{fig:Toy}(a) we have a toy 3D model with a coarse labeling: the normals directed to the right located on the left and, conversely, the normals directed to the left located on the right likely belongs to blue facets; both on the left and on the right  sides the normals directed upwards likely belong to red facets.
The term  $E_{\textrm{norm}}$ aims at collecting such likelihoods and extend them locally, so to obtain the result in Figure \ref{fig:Toy}(b). To compute the coarse labeling we first apply a simplified version of the MRF that minimizes the energy:
\begin{equation}
E_{\textrm{simpl}}(\mathbf{l}) =  \sum_{f} E_{\textrm{data}}(l_f) + \sum_{f,h} E_{\textrm{smooth}}(l_f) 
\end{equation}

\begin{figure}[t]
\centering
\includegraphics[width=0.45\textwidth]{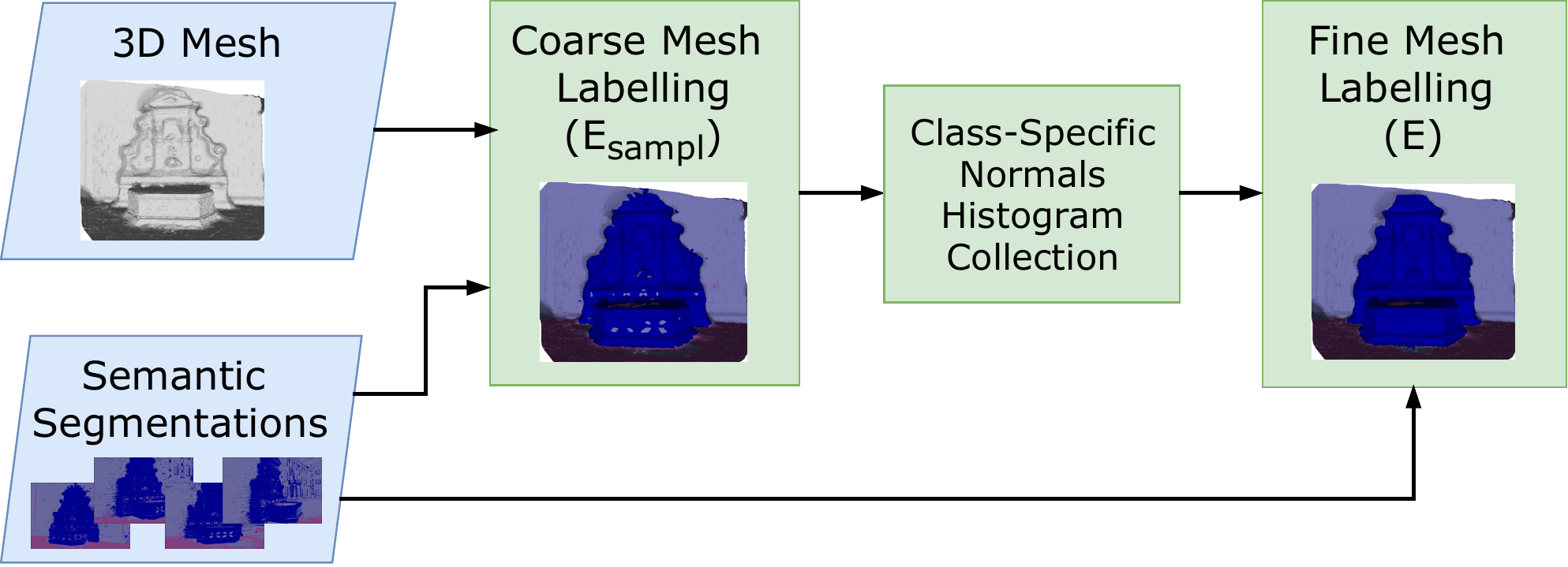}
\caption{Pipeline of the proposed labeling algorithm}
\label{fig:pipeline}
\end{figure}

\noindent Figure \ref{fig:pipeline}  summarizes the pipeline of our algorithm.

\subsection{Data and Smoothness Terms}
The term $E_{\textrm{data}}$ collects the evidences estimated by the image classifier among the different views. 
Since we know the likelihoods $I_i(l,x)$ for label $l$ in the pixel $x$ of the $i$-th image, we define:
\begin{equation}
    E_{\textrm{data}}(l_f) = -log\left( \sum_{i} \int_{\Omega_i}I_i(l_f,x)dx\right)
\end{equation}
where $\Omega_i$ is the area of the $i$-th image where the 3D model is projected.

We define the smoothness term between two adjacent facets as:
\begin{align}
E_{\textrm{smooth}}(l_f,l_h)= 
\begin{cases}
1.0, & \text{if } l_f \neq l_h\\
0.0, & \text{if } l_f = l_h
\end{cases}
\end{align}
to penalize frequent class changes among nearby facets.

\subsection{Discontinuity Prior}
Given two adjacent facet $f_1$ and $f_2$ with the corresponding normals $\mathbf{n}_{f_1}$ and $\mathbf{n}_{f_2}$, we define:

\begin{align}
E_{\textrm{disc}}(l_f,l_h) = 
\begin{cases}
e^{-\frac{(\angle(\mathbf{n}_{f_1},\mathbf{n}_{f_2}))^2}{2*(\frac{\pi}{2})^2}}, & \text{if } l_f \neq l_h\\
0.0, & \text{if } l_f = l_h
\end{cases}
\end{align}

This term represents the idea that two adjacent objects or two adjacent parts of the scene that belong to two different classes show discontinuities along the the mutual boundary. 
For instance, in the region between a car and the street the facets incident to the wheel-street boundary have significantly different orientations.
On the other hand is likely that adjacent facets with the same labels have similar orientation.
Only in the case of sharp edges the assumption does not hold, however even if we favor a label change in those areas, the data and smoothness terms provide enough evidence to avoid misclassification.

\begin{figure}[t]
\centering
\includegraphics[width=0.98\columnwidth]{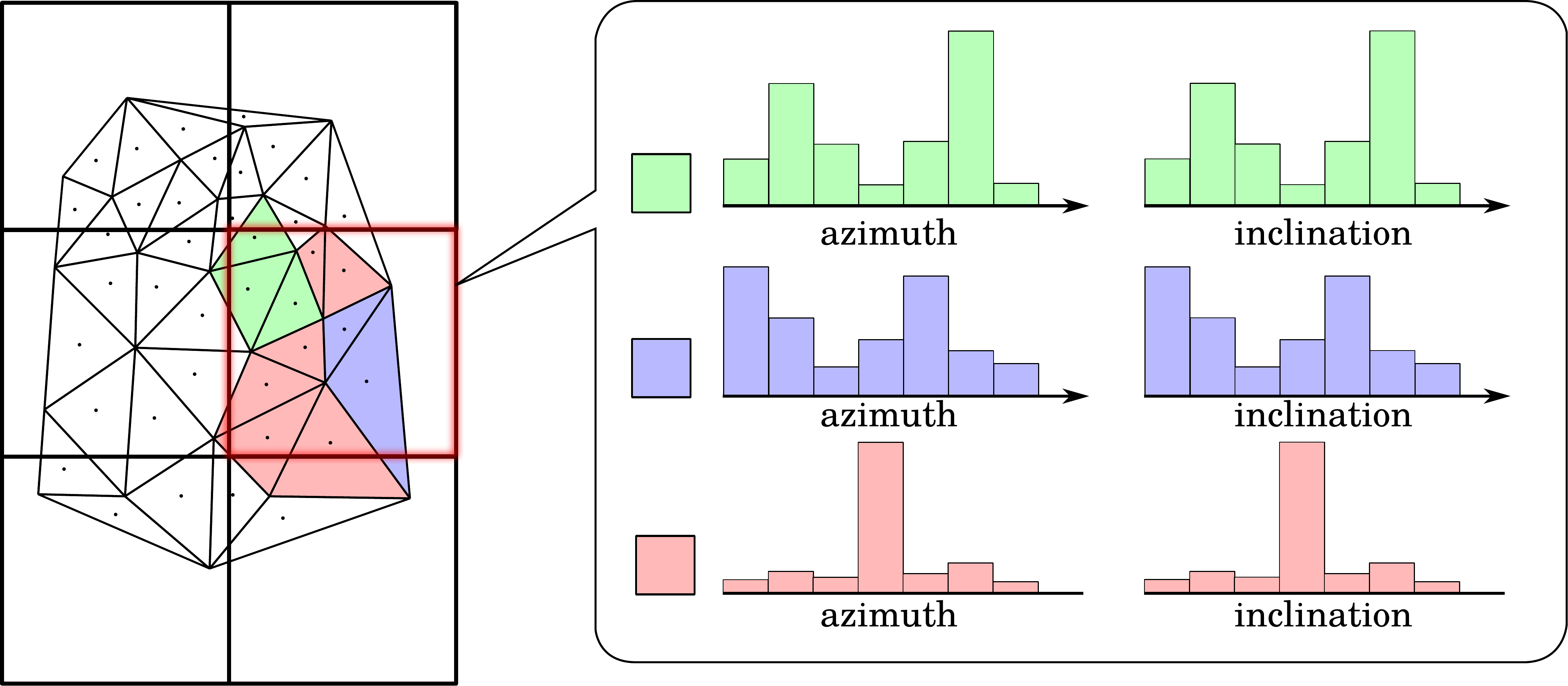}
\caption{Per-cell distributions of azimuth and inclination among the normal of the facts}
\label{fig:hists}
\end{figure}

\begin{table}[tp]
\small
\centering
\setlength{\tabcolsep}{2px}
  \caption{Resolutions and output statistic for each dataset for both the high and low resolution models}
  \label{tab:datasetStat}
\begin{tabular}{lccc}
&num.& image &num.\\
&cameras& resolution&facets  \\
\hline
fountain-P11    &11     & 3072x2048     & 1.9M\\
castle-P30      &30     & 3072x2048     & 5M\\
Southbuilding   &14     & 3072x2304     & 3M\\
KITTI 95        &548     & 1242x375     & 1.2M\\ 
Dagstuhl        &67     & 1600x1200     & 4.7M\\
\end{tabular}
\end{table}

\begin{figure}[tp]
\centering
\resizebox {0.7\columnwidth} {!} {
  \begin{tikzpicture}
  \begin{axis}[
      enlargelimits=false,
      xlabel={Voxel size (\% scene size)},
      ylabel={Intersection over Union},
       xmin=10, xmax=100,
       ymin=0.90, ymax=0.97,
      xtick={10,20,30,40,50,60,70,80,90,100},
      ytick={0.90,0.91,0.92,0.93,0.94,0.95,0.96,0.97},
      legend pos=north east,
      ymajorgrids=true,
      grid style=dashed,
  ]
  \addplot[
      color=blue,
      mark=none]
  table[x index=0,y index=2,col sep=comma]
  {data/varying_castle_IOU.txt};
  \addplot[
      color=red,
      mark=none]
  table[x index=0,y index=3,col sep=comma]
  {data/varying_castle_IOU.txt};
  \addplot[
      color=green,
      mark=square]
  table[x index=0,y index=1,col sep=comma]
  {data/varying_castle_IOU.txt};
  \legend{\cite{blaha2017semantically},\cite{romanoni2017multi},Proposed}
  \end{axis}
  \end{tikzpicture}
}
\caption{IoU with varying voxel size}
\label{fig:iouVoxel}
\end{figure}
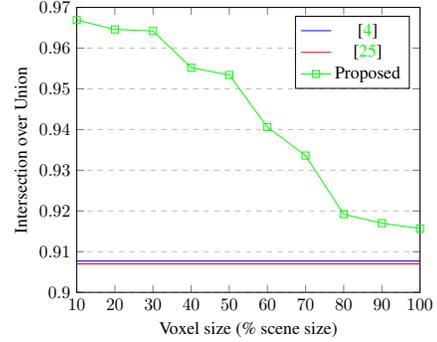

\begin{table*}[t]
\centering
 \footnotesize
  \caption{Segmentation results w.r.t. the state of the art and the baseline method without the proposed term $E_{\textrm{norm}}$}
  \label{tab:resTot}
    \begin{tabular}{llccccccccc}
dataset & method& average & average & average & average & overall  &  overall  &  overall   & overall &  IoU\\
      & & accuracy  &  recall  &  F-score   & precision & accuracy  &  recall   &  F-score   & precision &  \\
\hline 
\multirow{3}{*}{fountain-P11}
& Blaha \etal~\cite{blaha2017semantically}       & 0.9499 & 0.8468 & 0.9291 & 0.8776 & 0.9478 & 0.9132 & 0.9157 & 0.9110  & 0.8827\\
& Romanoni \etal~\cite{romanoni2017multi}     & 0.9476 & 0.8447 & 0.9291 & 0.8759 & 0.9449 & 0.9099 & 0.9115 & 0.9070  & 0.8793\\
& Baseline (w/o $E_{\textrm{norm}}$) & 0.9481  &  0.8389 &   0.9244  &  0.8701   & 0.9464  &  0.9118  &  0.9124   & 0.9082    &    0.8800  \\
& Proposed  & \textbf{0.9548} & \textbf{0.8536} & \textbf{0.9384} & \textbf{0.8851} & \textbf{0.9540} & \textbf{0.9228} & \textbf{0.9248} & \textbf{0.9202}  & \textbf{0.8900}\\
\hline 
\multirow{3}{*}{castle-P30} 
& Blaha \etal~\cite{blaha2017semantically}       & 0.9623 &          0.8401 & 0.7859 & 0.8114 & 0.9399 & 0.9757 & 0.8985 & 0.9352 & 0.9070\\
& Romanoni \etal~\cite{romanoni2017multi}    & 0.9628 &          \textbf{0.8986} &  {0.8819} & {0.8886} & 0.9282 & \textbf{0.9798} & 0.8993 & 0.9366 &  0.9077\\
& Baseline (w/o $E_{\textrm{norm}}$)  & 0.9630 &          0.8666  &  0.7955 &   0.8264  &  {0.9410} &  0.9754 &   0.8995  &  0.9351           &    0.9079\\
&Proposed   & \textbf{0.9816}   & {0.8695} &\textbf{ 0.9423 }& \textbf{0.9008} & \textbf{0.9721} & 0.9774 &\textbf{ 0.9613} & \textbf{0.9686} & \textbf{0.9646}\\
\hline
\multirow{3}{*}{Southbuilding}
& Blaha \etal~\cite{blaha2017semantically}       & 0.9561 & \textbf{0.7717} & \textbf{0.8658} & \textbf{0.8129} & \textbf{0.9527} & 0.9040 & 0.9428 & 0.9225 & 0.8920\\
& Romanoni \etal~\cite{romanoni2017multi}  & 0.9507 & 0.7644 & 0.8419 & 0.8000 & 0.9444 & 0.8935 & 0.9328 & 0.9124 & 0.8839\\
& Baseline (w/o $E_{\textrm{norm}}$)  & 0.9097 &   0.6581  &  0.7333  &  0.6884  &  0.8763  &  0.8564  &  0.8398  &  0.8461    & 0.7984\\
&Proposed & \textbf{0.9572} & 0.7701 & 0.8651 & \textbf{0.8128} & \textbf{0.9523} & \textbf{0.9073} & \textbf{0.9448} & \textbf{0.9249} &  \textbf{0.8936}\\
\hline 
\multirow{3}{*}{KITTI 95}
& Blaha \etal~\cite{blaha2017semantically}      &  \textbf{0.9648} & 0.8601 & 0.8541 & 0.8588 & \textbf{0.9645} & 0.9155 & 0.8409 & 0.8744  & \textbf{0.8942} \\
&Romanoni \etal~\cite{romanoni2017multi}      &  0.9491 & 0.8006 & 0.7832 & 0.8232 & 0.9519 & 0.9004 & 0.8050 & 0.8445  & 0.8470 \\
& Baseline (w/o $E_{\textrm{norm}}$)  &  \textbf{0.9647} & 0.8599 & 0.8534 & 0.8579 & \textbf{0.9642} & 0.9158 & 0.8393 & 0.8736  & 0.8938 \\
& Proposed  &  \textbf{0.9648} & \textbf{0.8761} & \textbf{0.8682} & \textbf{0.8724} & \textbf{0.9642} & \textbf{0.9290} & \textbf{0.8509} & \textbf{0.8860}  & \textbf{0.8942} \\
\hline 
\multirow{3}{*}{Dagstuhl}
& Blaha \etal~\cite{blaha2017semantically}     &  0.9813 & \textbf{0.9291} & 0.9431 & 0.9581 & 0.9776 & \textbf{0.9258} & 0.9400 & 0.9327  & 0.9536 \\
& Romanoni \etal~\cite{romanoni2017multi}      &  0.9586 & 0.8327 & 0.8868 & 0.9519 & 0.9443 & 0.7936 & 0.9329 & 0.8561  & 0.8972 \\
& Baseline (w/o $E_{\textrm{norm}}$)  &  0.9592 & 0.8130 & 0.8856 & \textbf{0.9769} & 0.9427 & 0.7740 & \textbf{0.9701} & 0.8600  & 0.8839 \\
& Proposed  &  \textbf{0.9821} & 0.9241 & \textbf{0.9472} & 0.9717 & \textbf{0.9787} & 0.9192 & 0.9601 & \textbf{0.9391}  & \textbf{0.9561} \\
    \end{tabular}
\end{table*}

\subsection{Normal Likelihood}
The previous terms are able to map the label from 2D segmentation to the 3D mesh. 
However, in some cases the smoothing term is not able to tackle the noise affecting the segmentations.
In such cases the common approach is to define class specific priors as in~\cite{blaha2017semantically} that favor one class with respect to others depending on the surface orientation.
These priors are effective and they are able to compensate noise and classification errors; however, they require both the knowledge of the gravity vector and to define and tune, for each class, which orientations are more likely.

We replace the prior term, such that we no longer require prior knowledge about the environment, but we still favor a labeling depending on facets orientation. 
The result is a more flexible and general algorithm. 
We follow and extend the intuition reported in~\cite{romanoni2017multi}, that is, given a 3D mesh with a coarse labeling, we can collect, for each label, the distribution of the normals and use them to estimate the likelihood of a facet $f$ with a normal $n_f$ to belong to a class with label $l$, therefore we turn the prior term into a likelihood term.

For each class, the authors in~\cite{romanoni2017multi} \etal collect the mean normal and its variance: if the facets belonging to a class of label $l$, \eg, the ground, show a strong tendency towards a single orientation, \ie, the normals have a unimodal distribution with low variance, the method favors the other facets with similar orientation to be labeled as $l$.
This ``global'' approach fails in two cases. 
One case is exemplified as follows: if the algorithm biases all the facets oriented upwards to become ground, also the roofs of the cars, or some roofs of buildings are biased to be ground. 
The second case happens if the normals (of the facets) belonging to a class have a multimodal distribution, such as the walls in a street; in this scenario the single mean and variance computed in~\cite{romanoni2017multi} are not expressive enough.

To face these two issues, we propose to locally estimate  the distributions of the normals for each class/label, and to represent and use directly these distributions as a likelihood term in the energy function instead of defining it from the mean and the variance.

First we define a 3D lattice that covers the whole 3D mesh, its cells are cubes of size $d_i$, \eg, the left part of Figure \ref{fig:hists} exemplifies it in 2D. 
For each cell $c_k$ and each class $l$ we turn the normal of the facets whose centroid is inside of $c_i$ into unit spherical coordinates.
Then, we define two histograms $hist_{k,l}^{\text{azim}}$ and  $hist_{k,l}^{\text{incl}}$ to collect respectively the distribution of the normal azimuths and inclinations.

When we evaluate a facet $f$ with the centroid inside the cell $c_k$, we define the likelihood:
\begin{equation}
    P_k (f|l) \sim e^{\text{hist}_{k,l}^{\text{azim}}(\text{az}(\mathbf{n}_f)) * \text{hist}_{k,l}^{\text{incl}}(\text{inc}(\mathbf{n}_f))}
\end{equation}
where $\text{az}(\cdot)$ and $\text{inc}(\cdot)$ convert a vector into its azimuth and inclination.
Therefore the normal likelihood term is:
\begin{equation}
    E_{\textrm{norm}} = -log(\text{hist}_{k,l}^{\text{azim}}(\text{az}(\mathbf{n}_f)) * \text{hist}_{k,l}^{\text{incl}}(\text{inc}(\mathbf{n}_f)))
\end{equation}

\paragraph{Normal Likelihood vs. Smoothing}
The term $E_{\text{norm}}$ let that, locally, facets with similar orientation takes similar labels. 
Even if this behavior may seems similar to a classical contrast-sensitive smoothing term, its meaning, implementation and result is different.
The two-step approach (coarse labeling with histogram computation + fine labeling) let us to define $E_{\text{norm}}$ as a simple unary term for each facet, instead of a pairwise term, as the usual smoothing term. 
This means that during the optimization procedure $E_{\text{norm}}$ does not change its value for each facet $f$, similarly to what happens with prior terms differently from the behavior of a smoothness term that spreads the label with current higher support.
In principle, our method could be imitated by a classic smoothing term that adds a pairwise term connecting every facet to all the facets inside the big voxel of the grid. However in this case the dimension of the MRF would quickly become intractable, \eg, in the fountain-P11 dataset we have around 12000 facets for each voxel, therefore this approach would require $\frac{12000!}{2!(2-12000)!}\approx72\cdot10^{6}$ pairwise terms for each facet.

\section{Experiments}
\label{sec:exp}
We tested our labeling method against five datasets: fountain-P11, castle-P30~\cite{strecha2008}, Southbuilding~\cite{hane2013joint}, KITTI 95~\cite{geiger_et_al12} and Dagstuhl.
In Table \ref{tab:datasetStat} we illustrate the statistics of the datasets and the mesh we estimated.  
Even though it is related to the problem we faced in this paper, we did not used the Varcity dataset~\cite{riemenschneider2014learning}, indeed the classes are all spread along the building facade, and in this case a class-specific normal prior does not have any influence on the labeling stage.
To extract the input mesh we calibrate the cameras with Structure from Motion (SfM)~\cite{openMVG}, and we estimate the depth maps with the plane sweeping library~\cite{hane2014real,PSL}.
We fuse SfM points and the depth maps by means of the volumetric reconstruction algorithm proposed in~\cite{romanoni15b}, and we refined the model through photometric refinement~\cite{vu_et_al_2012}.
For fountain-P11, castle-P30, Southbuilding and Dagstuhl datasets, we extract the 2D segmentation with the per-pixel likelihood score through the Multiboost classifier~\cite{benbouzid2012multiboost}: we manually labeled few images and we classify all the datasets images into \textit{ground}, \textit{wall}, \textit{vegetation} and \textit{other}. For the KITTI sequence we used a pre-trained version of~\cite{visin2016reseg}.
We modeled the MRF through the openGM library~\cite{opengm-library} and we adopted the alpha expansion algorithm~\cite{boykov2001fast} to infer the labeling.

\begin{figure*}[tpb]
\centering
\setlength{\tabcolsep}{1px}
\begin{tabular}{ccccc}
2D & GT & Blaha \etal & Romanoni \etal &Proposed\\
Segmentation& &\cite{blaha2017semantically}&\cite{romanoni2017multi}  &\\
\hline
\multicolumn{5}{c}{fountain-P11}\\
\includegraphics[width=0.18\textwidth]{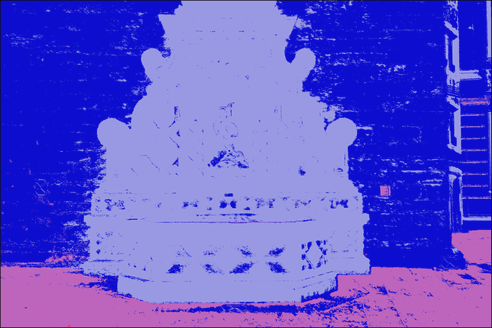}&
\includegraphics[width=0.18\textwidth]{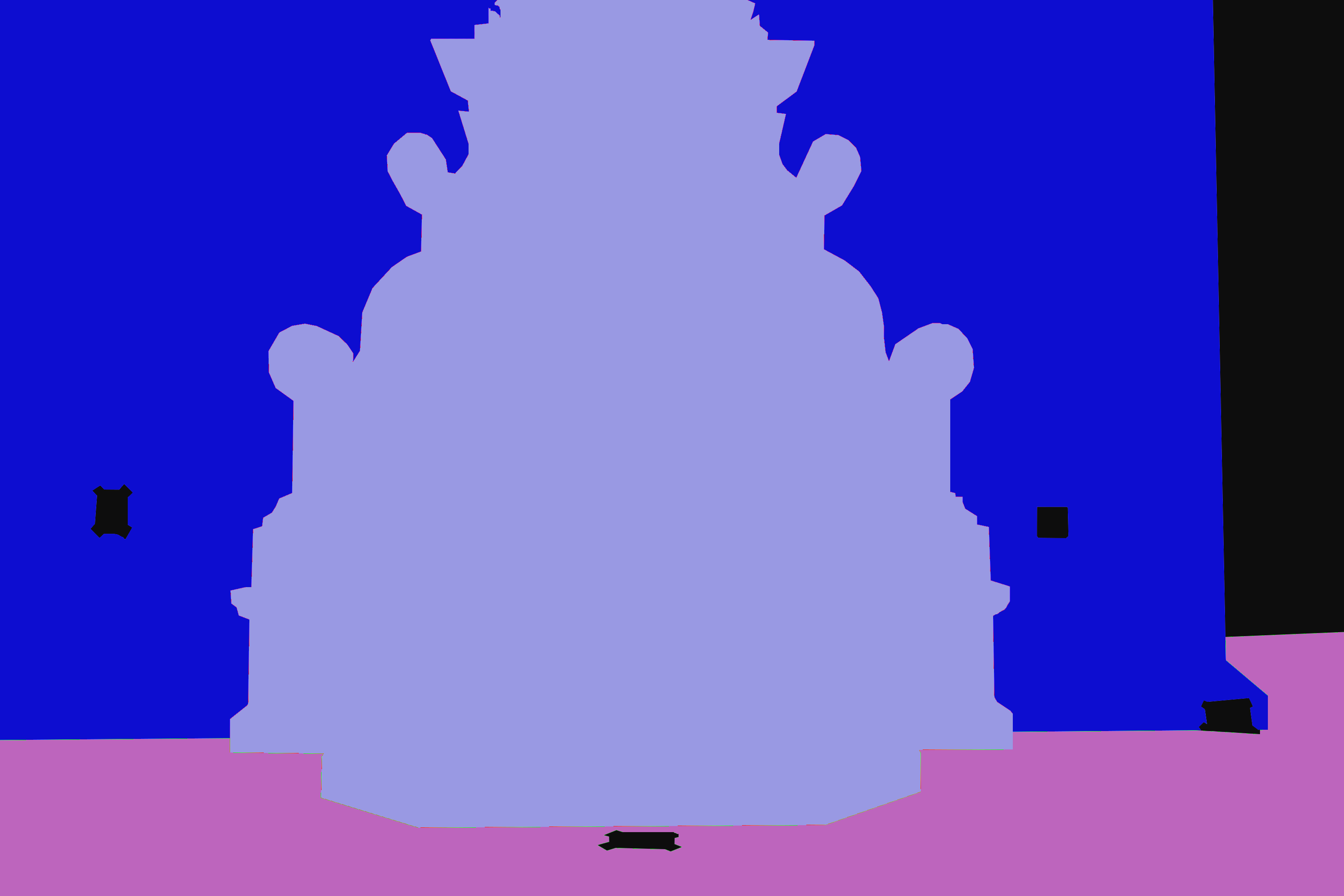}&
\includegraphics[width=0.18\textwidth]{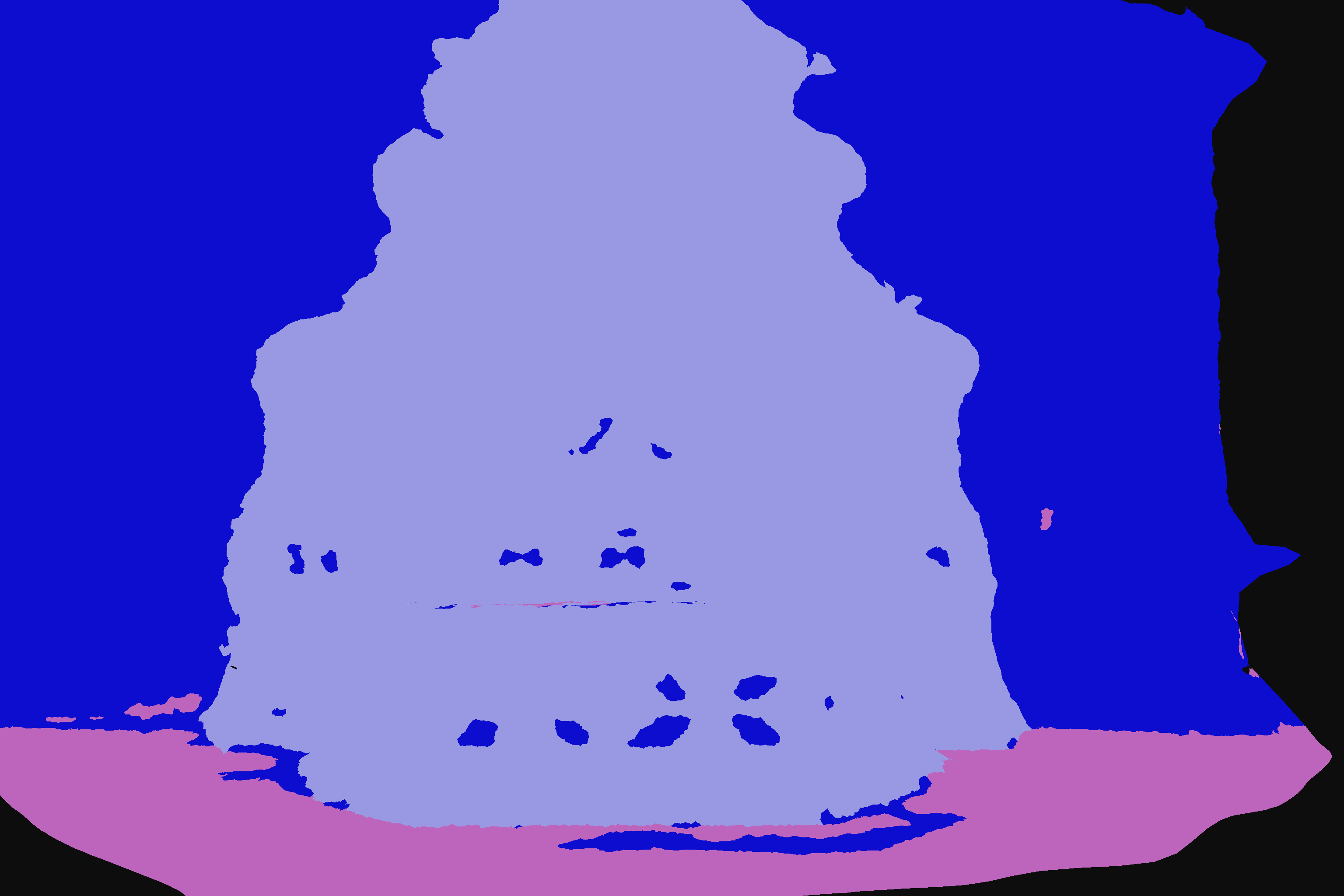}&
\includegraphics[width=0.18\textwidth]{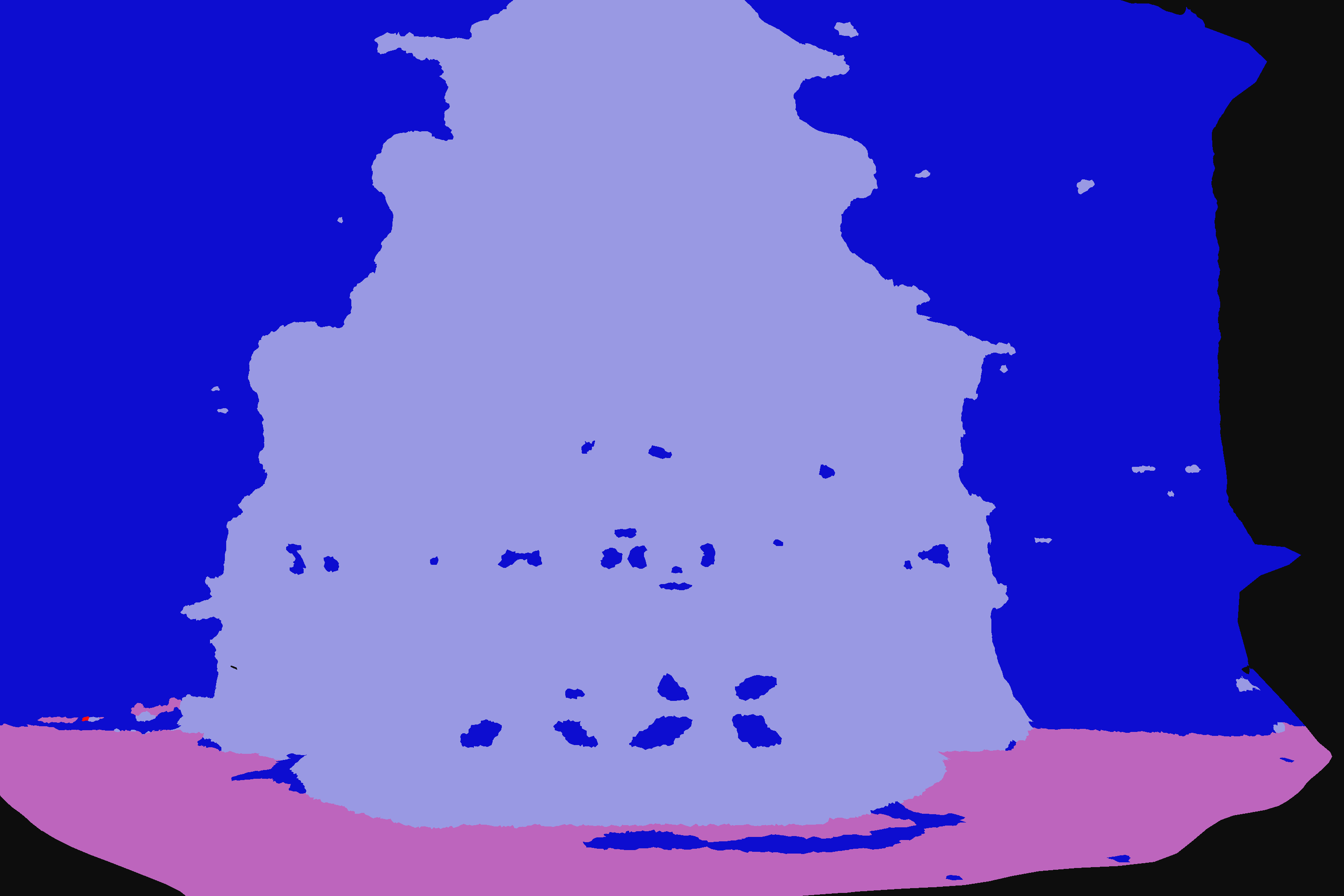}&
\includegraphics[width=0.18\textwidth]{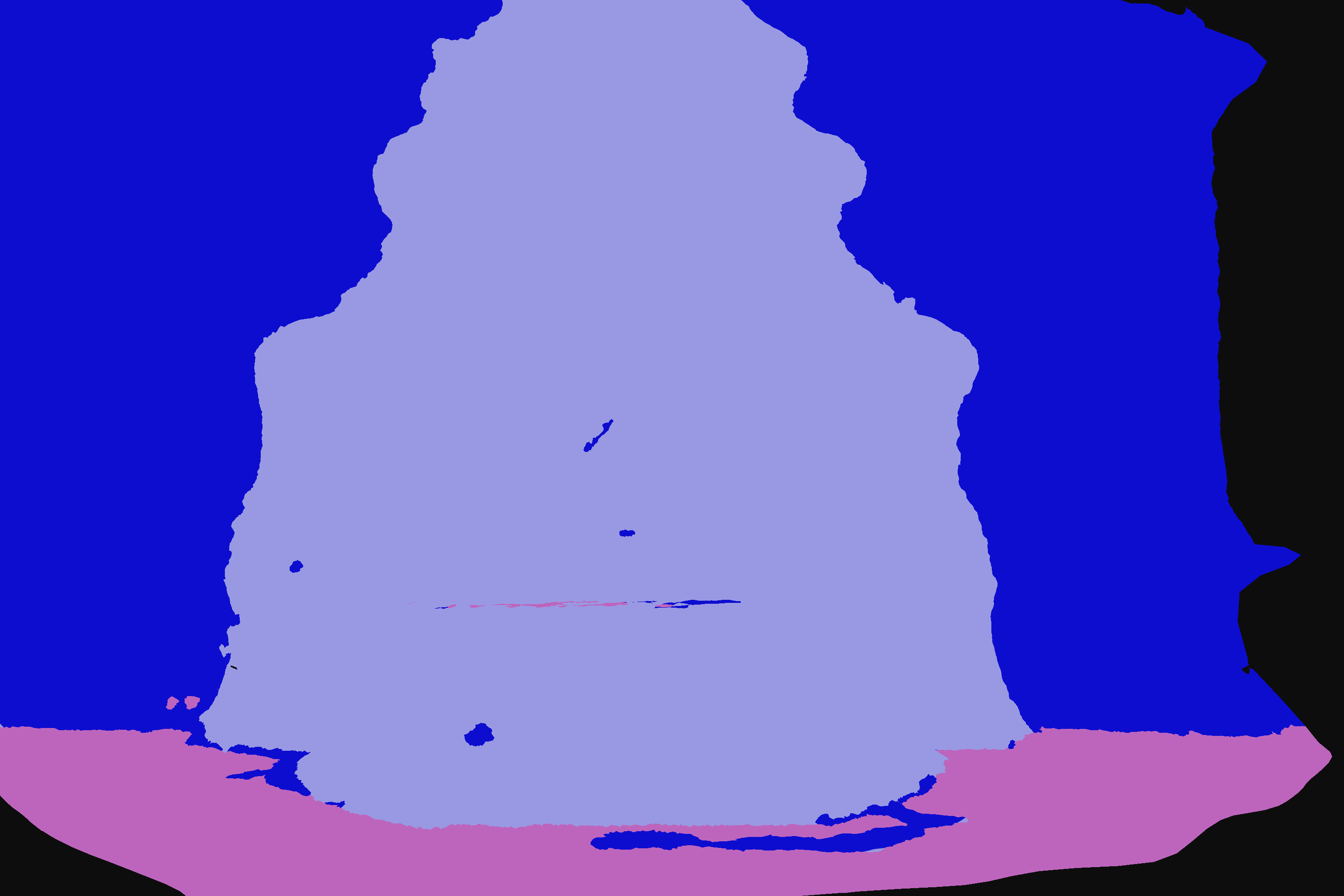}\\
\multicolumn{5}{c}{castle-P30}\\
\includegraphics[width=0.18\textwidth]{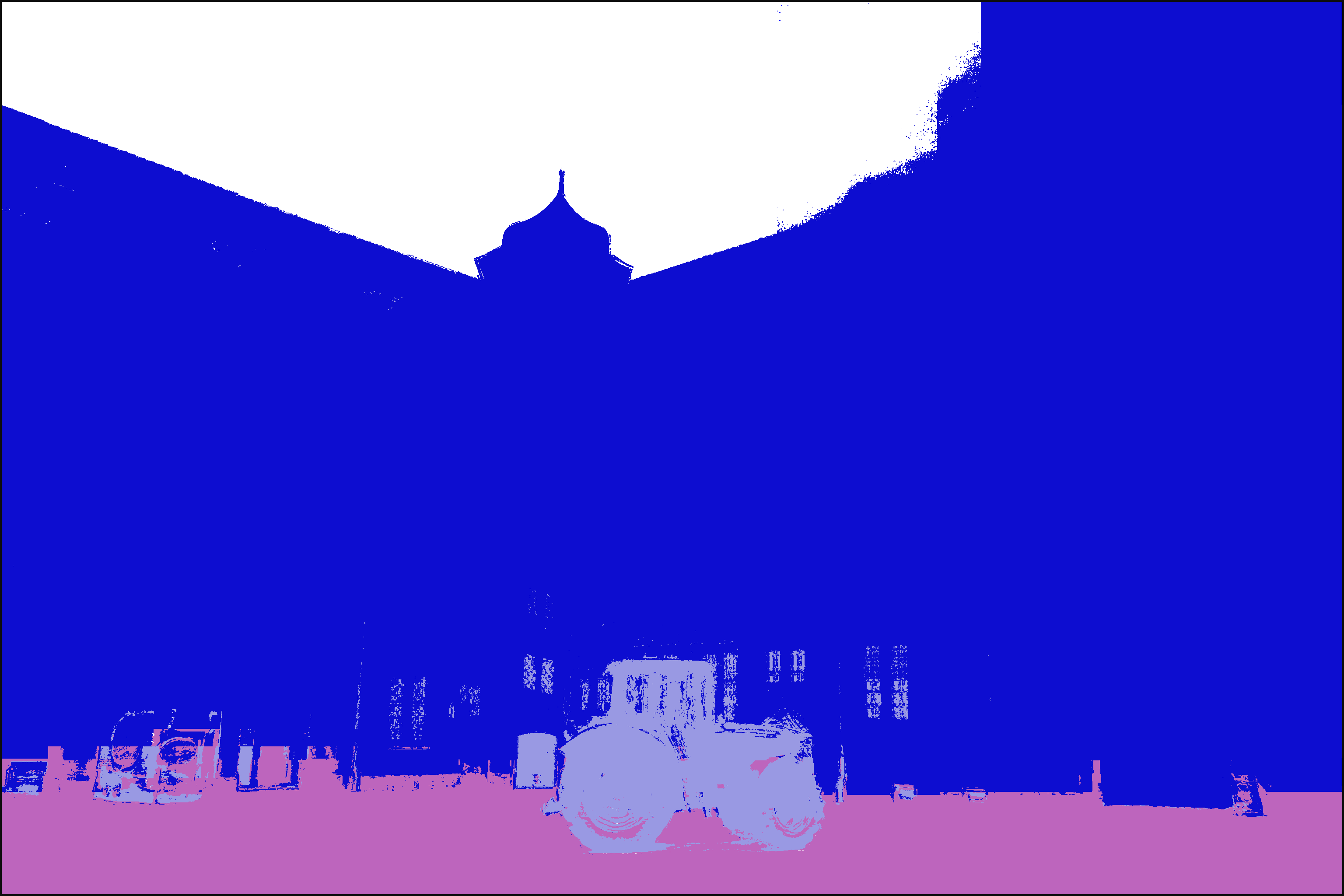}&
\includegraphics[width=0.18\textwidth]{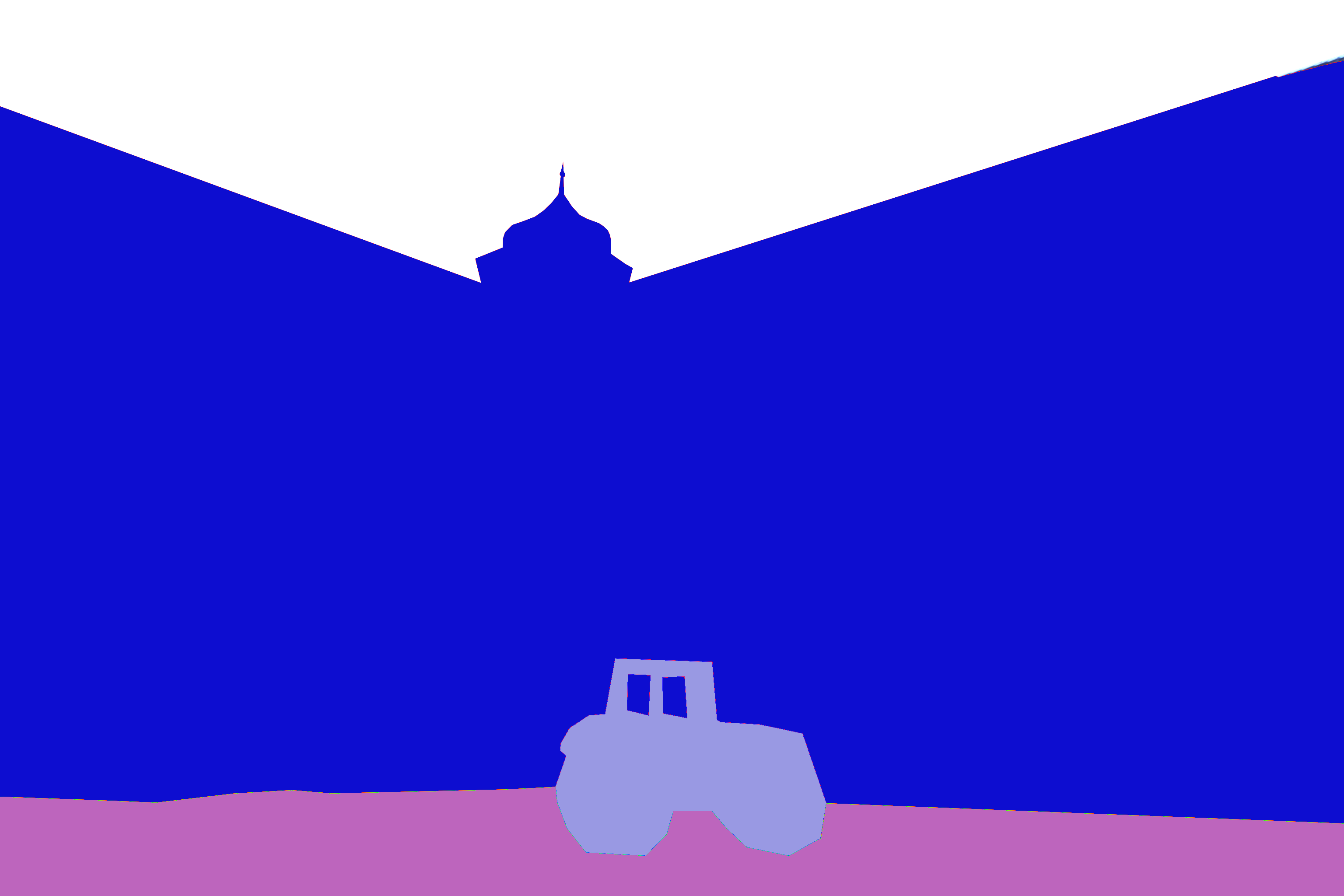}&
\includegraphics[width=0.18\textwidth]{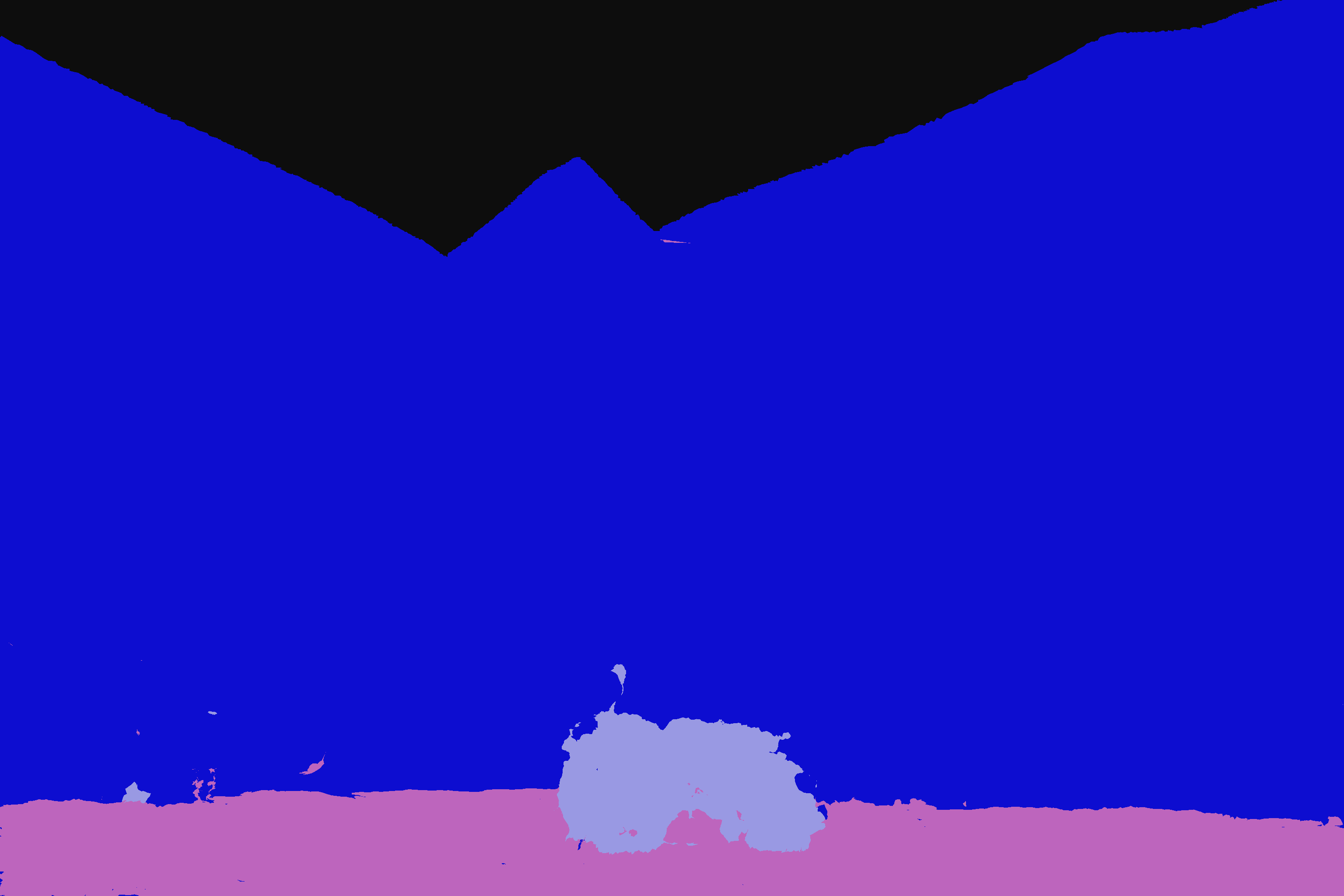}&
\includegraphics[width=0.18\textwidth]{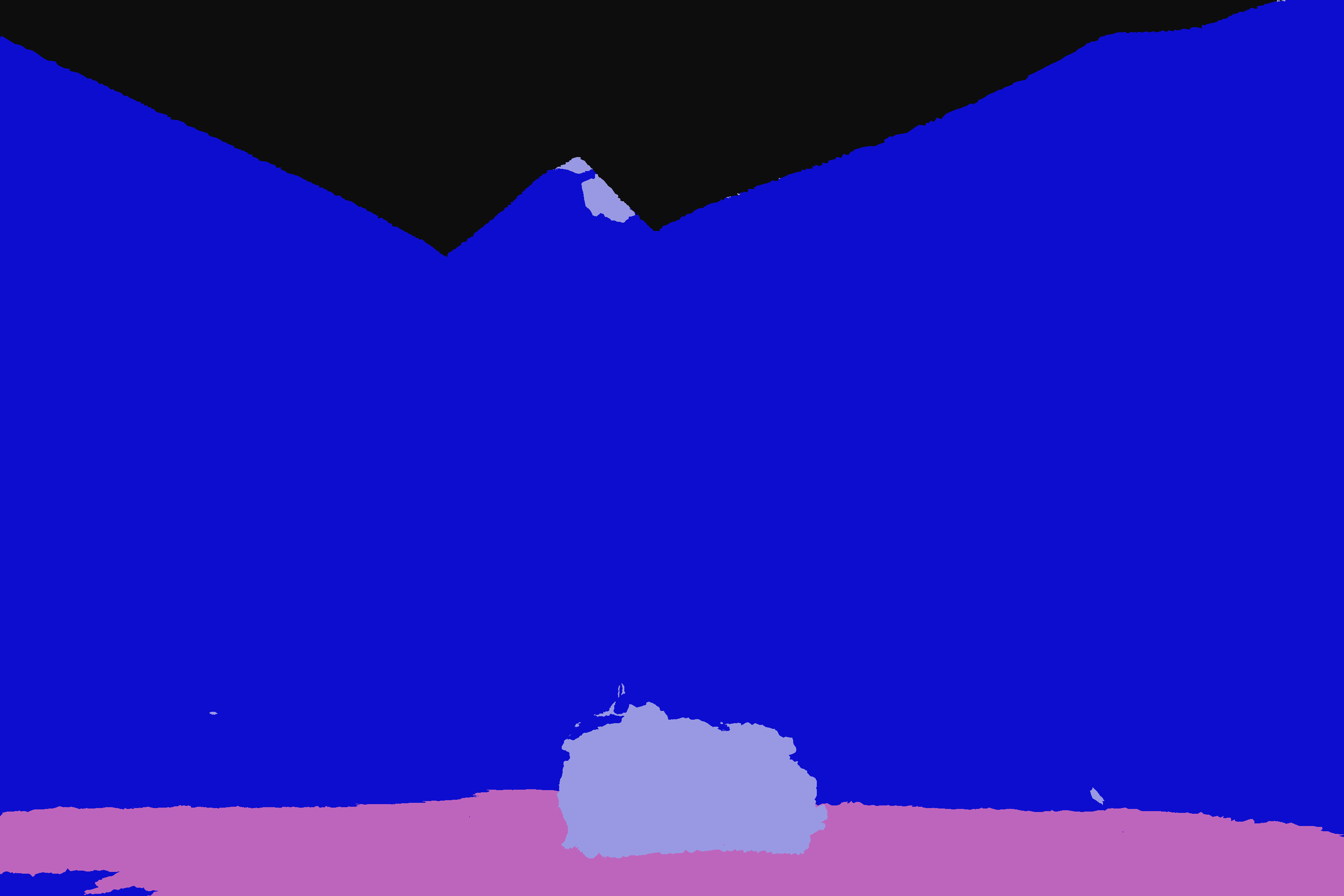}&
\includegraphics[width=0.18\textwidth]{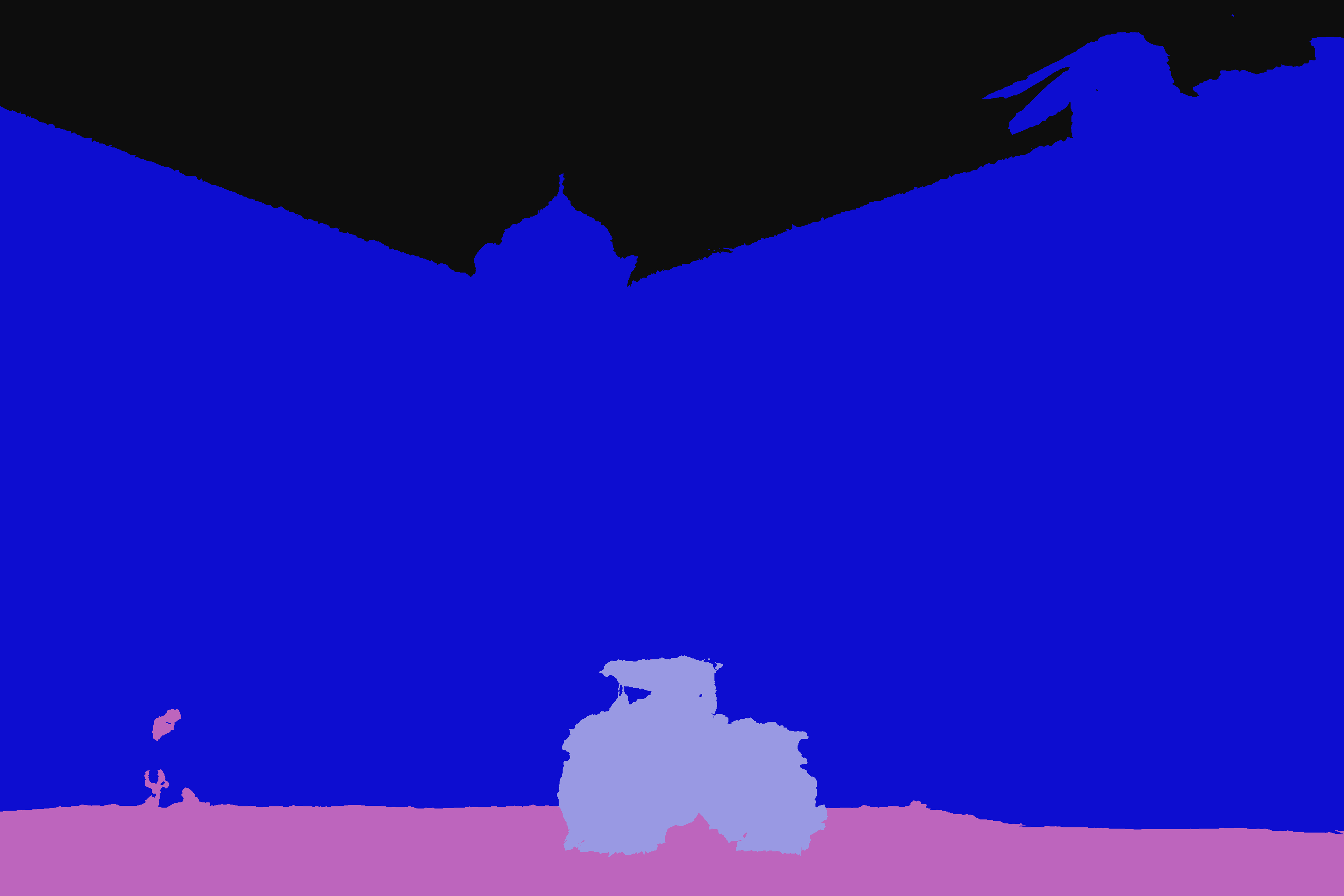}\\
\multicolumn{5}{c}{Southbuilding}\\
\includegraphics[width=0.18\textwidth,height=0.125\textwidth]{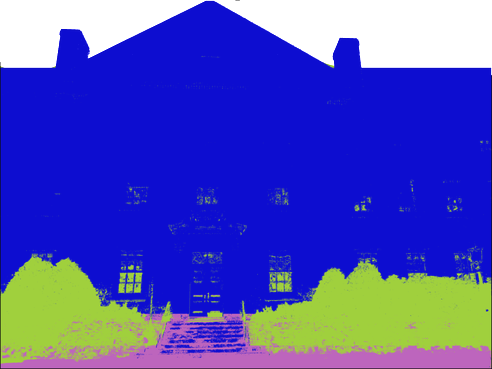}&
\includegraphics[width=0.18\textwidth,height=0.125\textwidth]{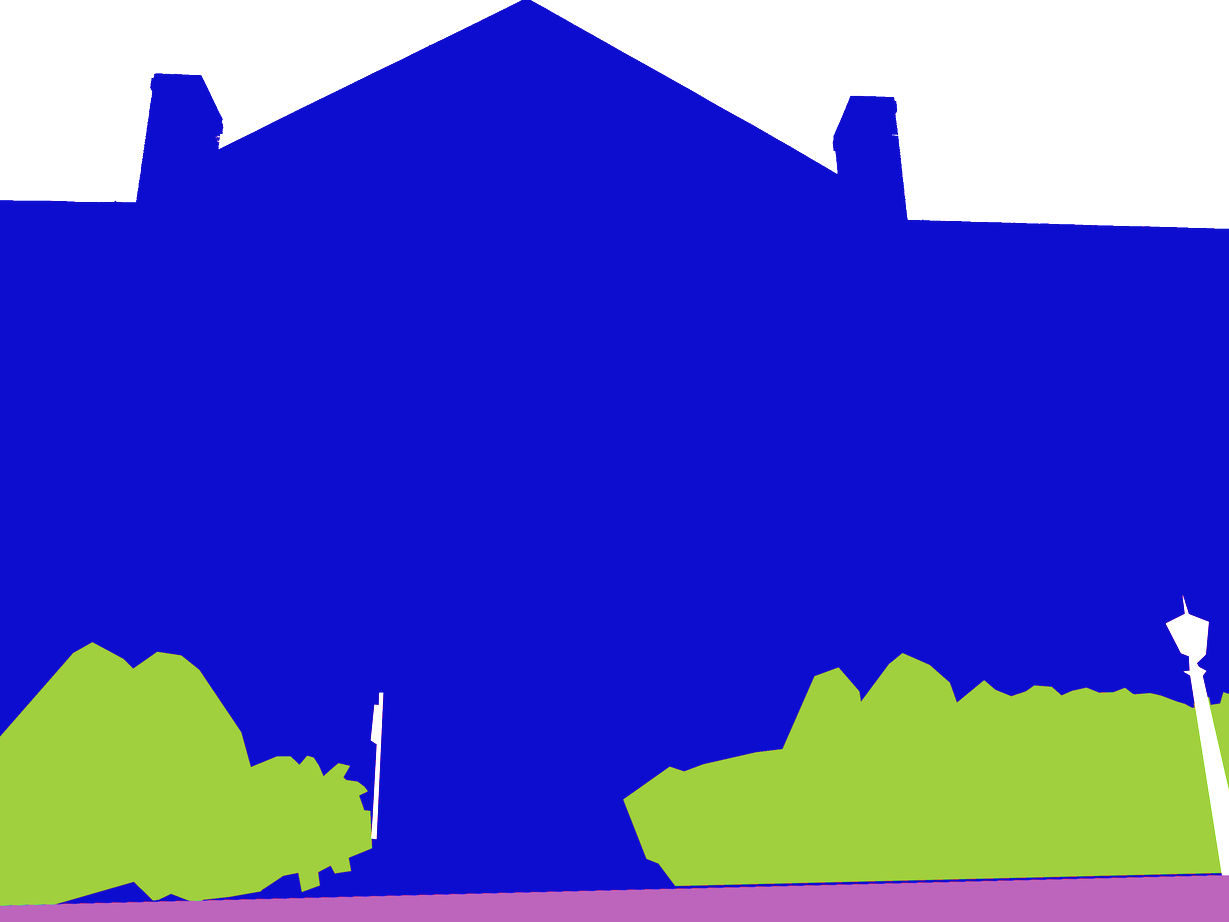}&
\includegraphics[width=0.18\textwidth,height=0.125\textwidth]{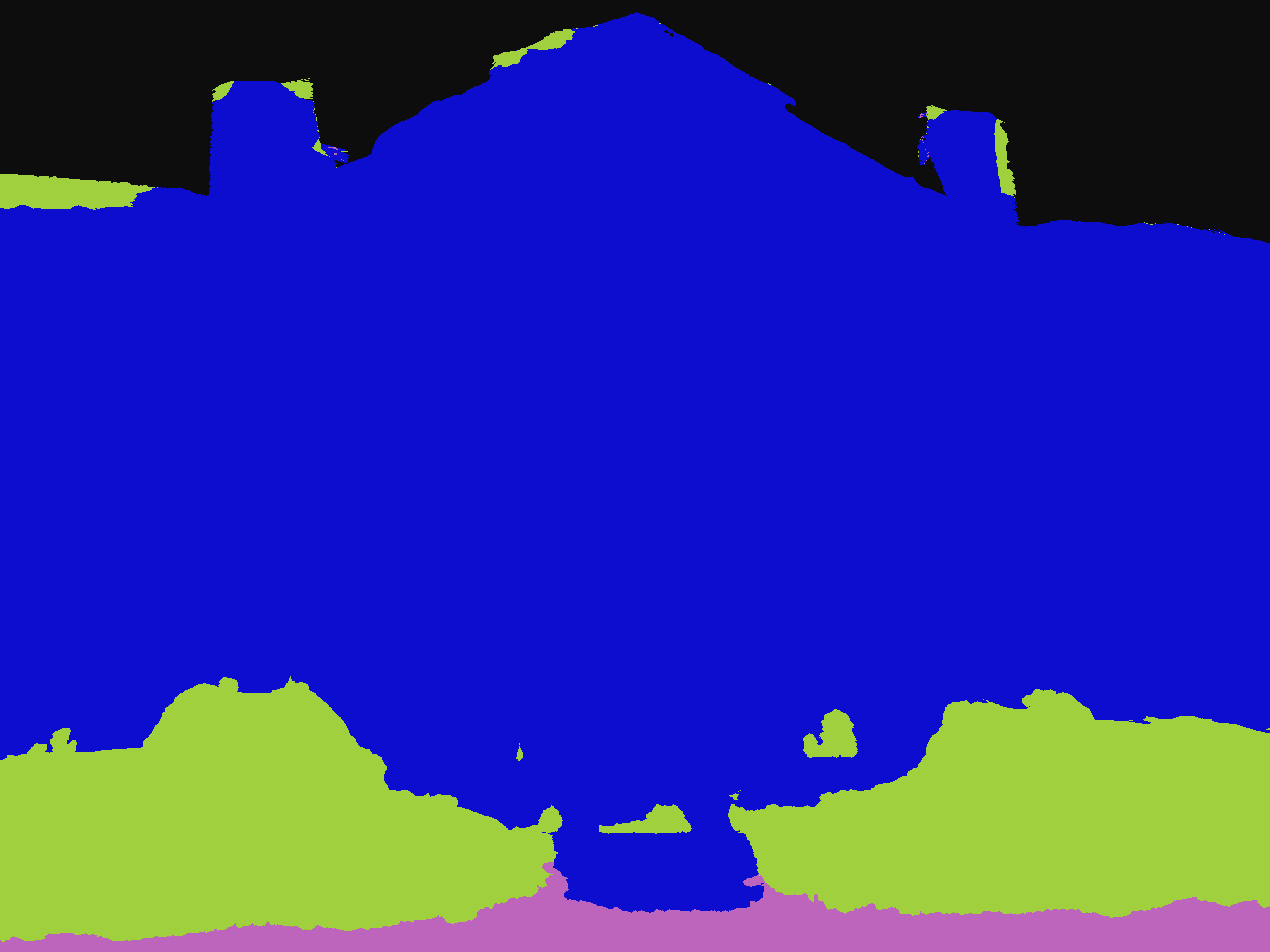}&
\includegraphics[width=0.18\textwidth,height=0.125\textwidth]{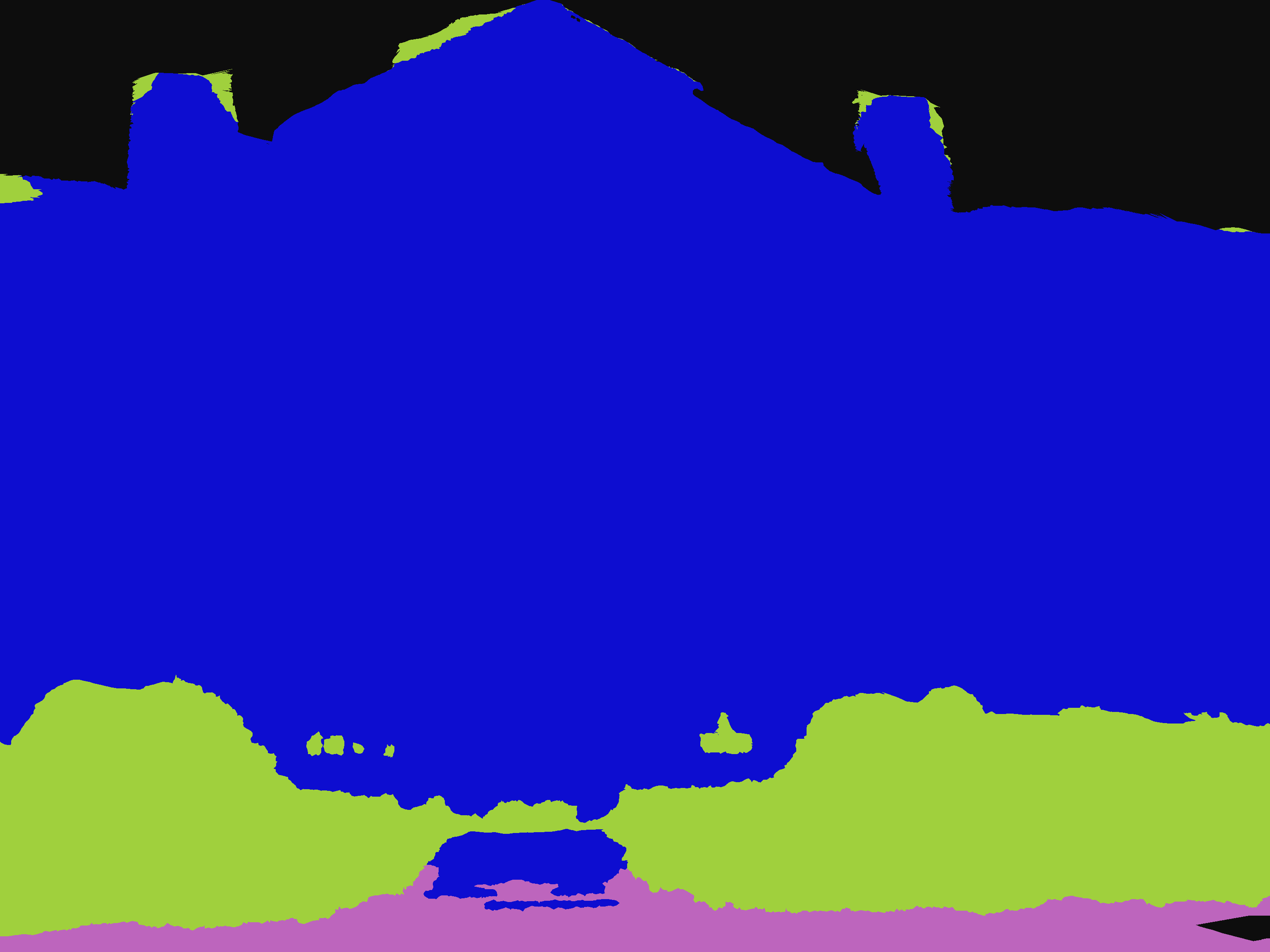}&
\includegraphics[width=0.18\textwidth,height=0.125\textwidth]{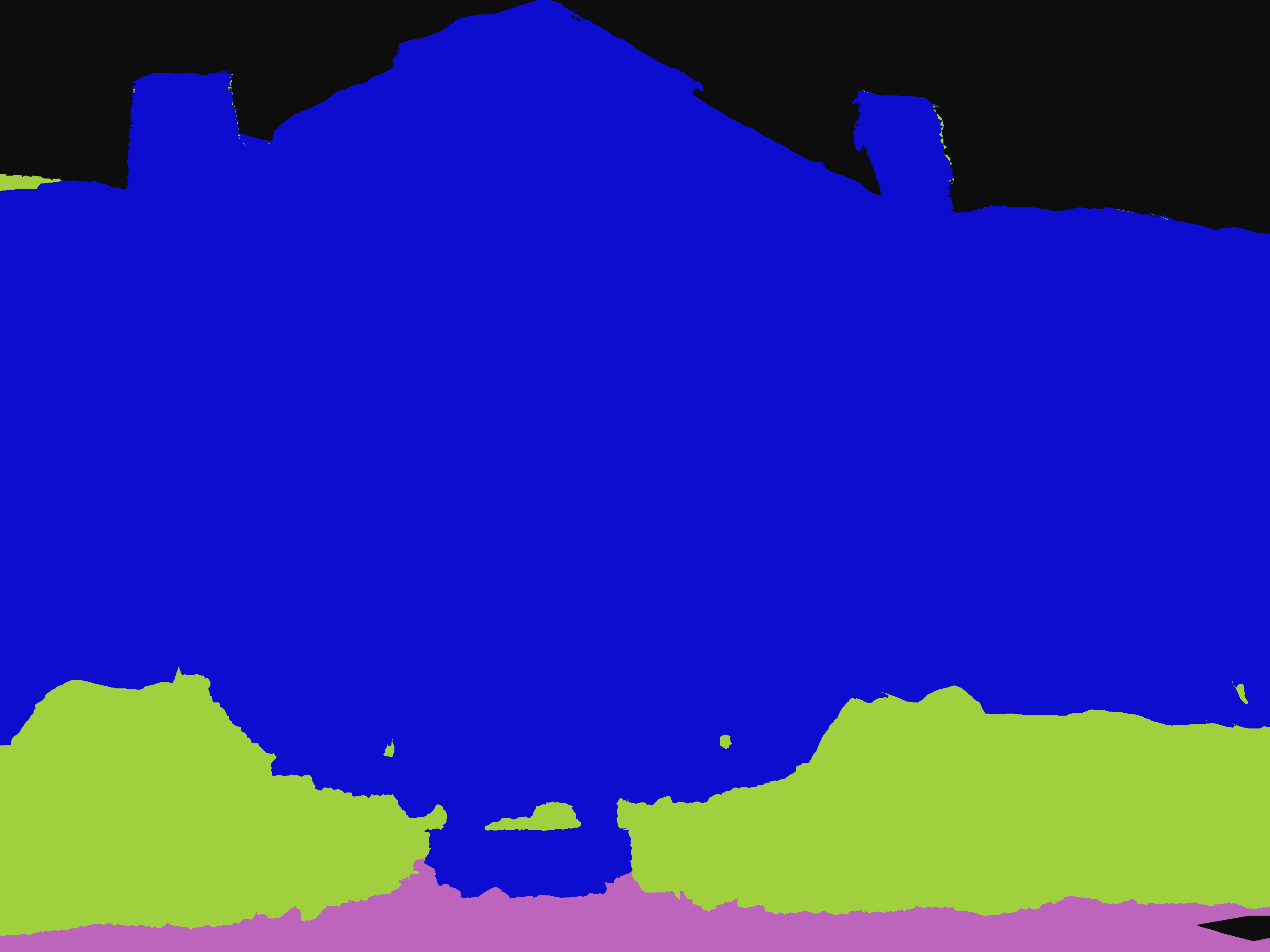}\\
\multicolumn{5}{c}{KITTI 95}\\
\includegraphics[width=0.18\textwidth,height=0.125\textwidth]{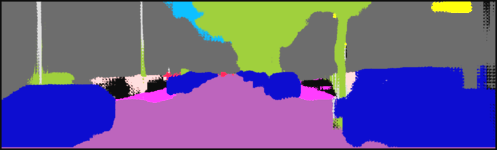}&
\includegraphics[width=0.18\textwidth,height=0.125\textwidth]{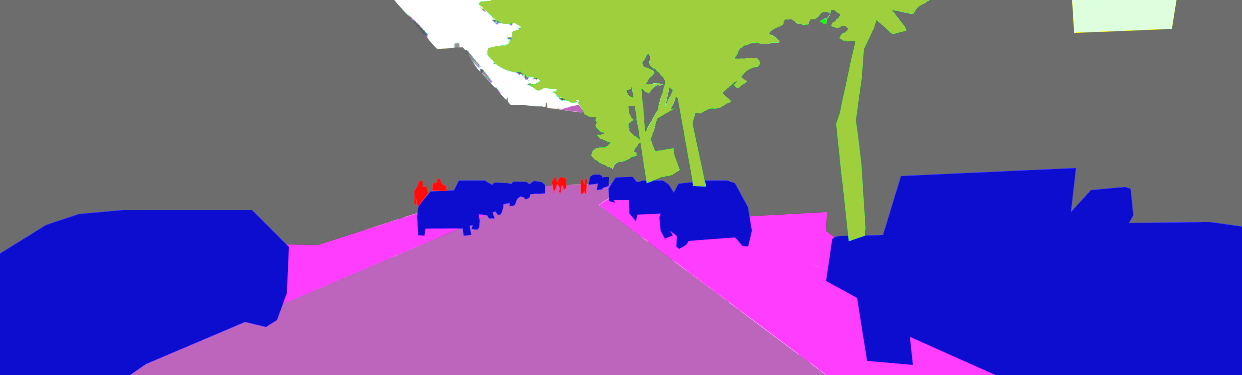}&
\includegraphics[width=0.18\textwidth,height=0.125\textwidth]{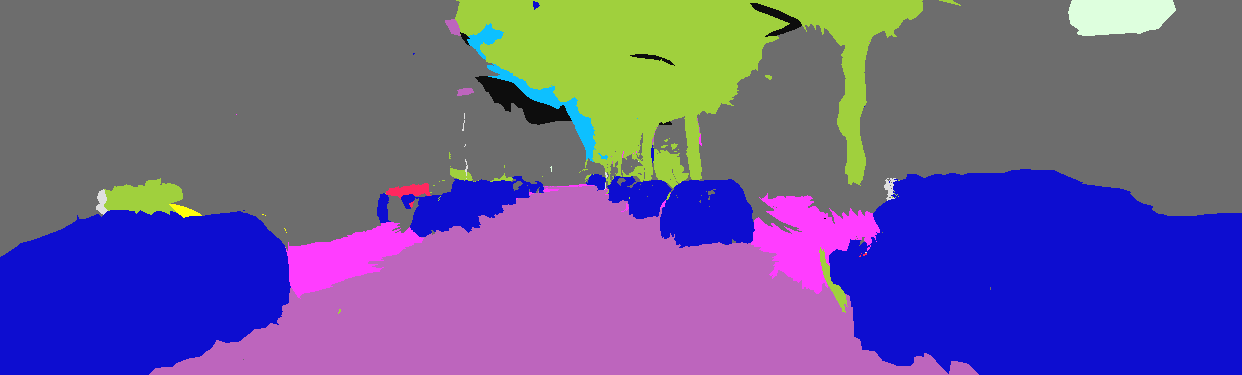}&
\includegraphics[width=0.18\textwidth,height=0.125\textwidth]{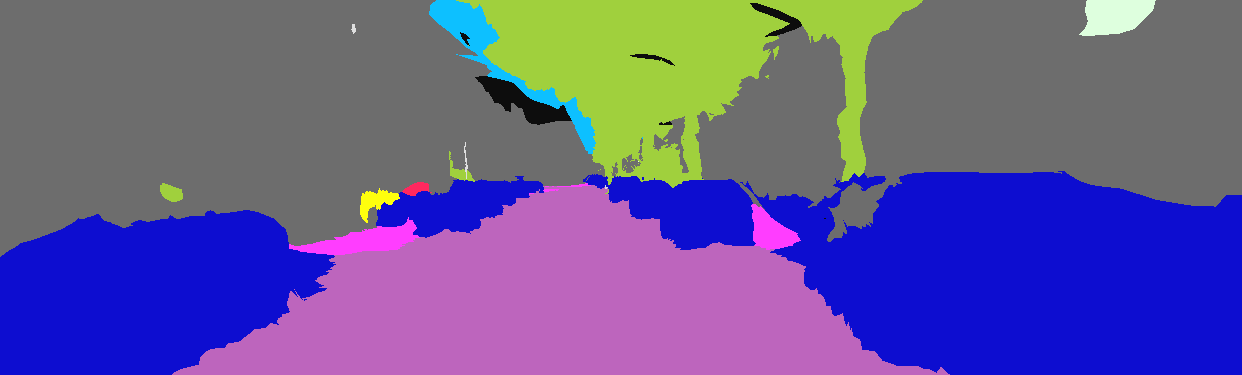}&
\includegraphics[width=0.18\textwidth,height=0.125\textwidth]{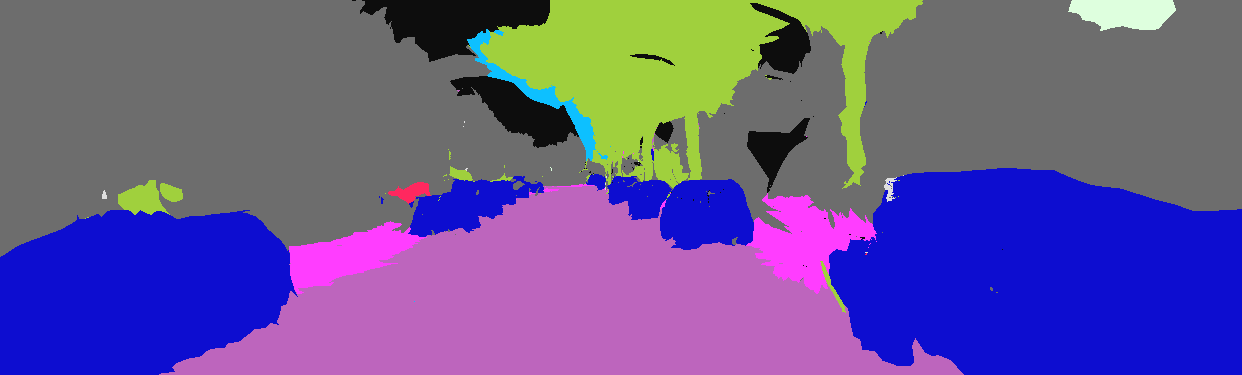}\\
\multicolumn{5}{c}{Dagstuhl}\\
\includegraphics[width=0.18\textwidth,height=0.125\textwidth]{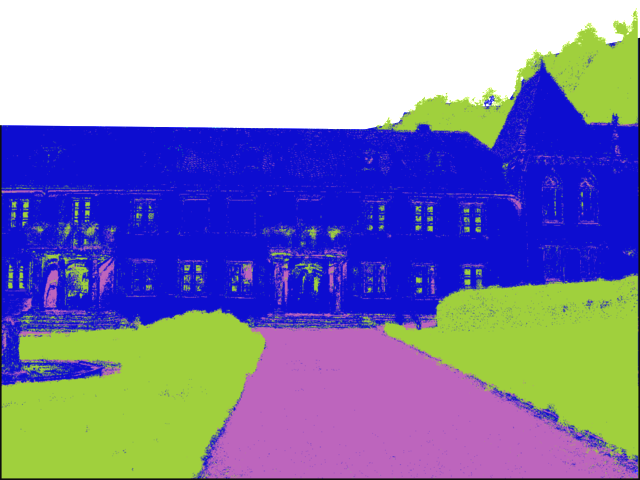}&
\includegraphics[width=0.18\textwidth,height=0.125\textwidth]{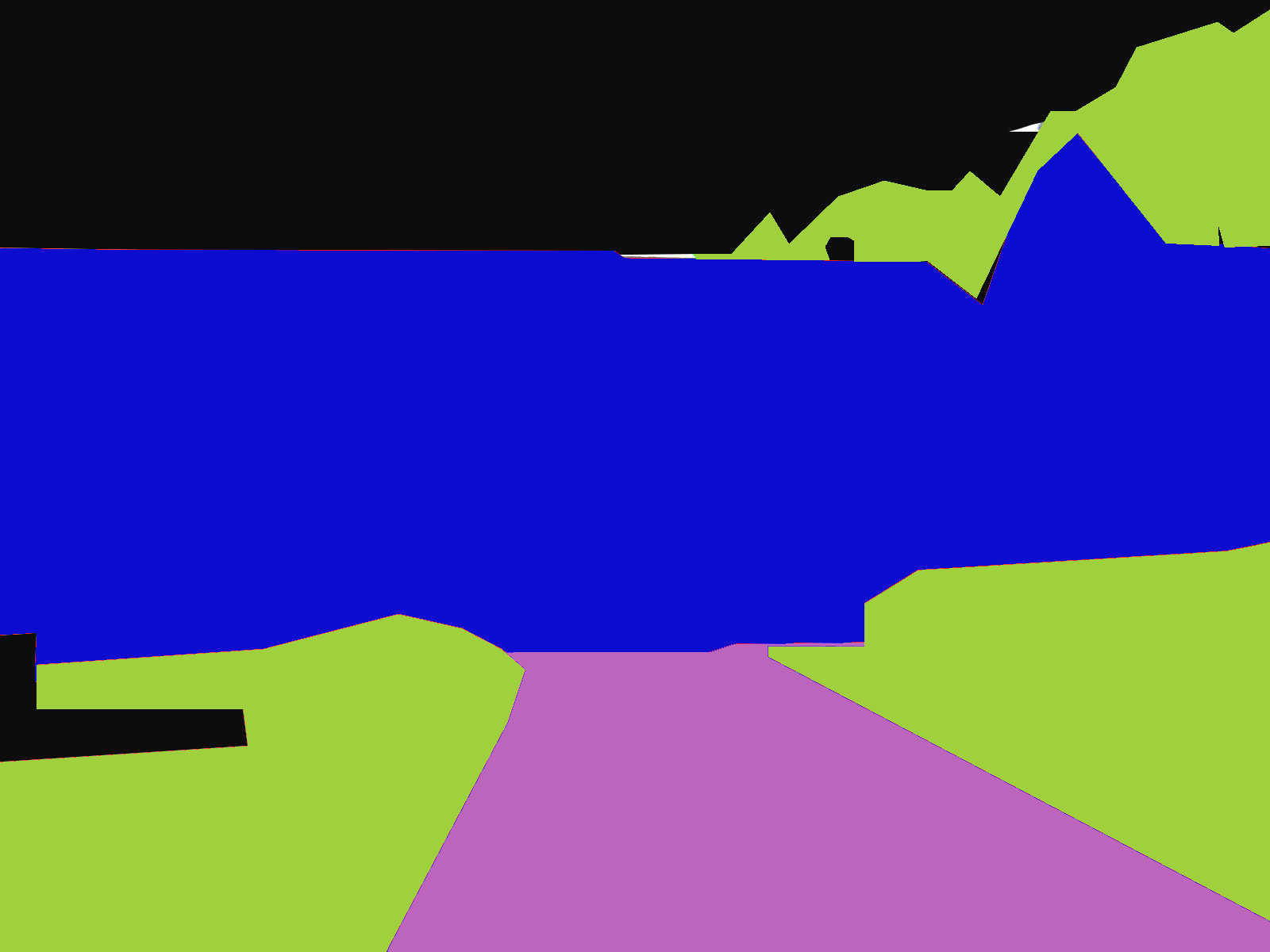}&
\includegraphics[width=0.18\textwidth,height=0.125\textwidth]{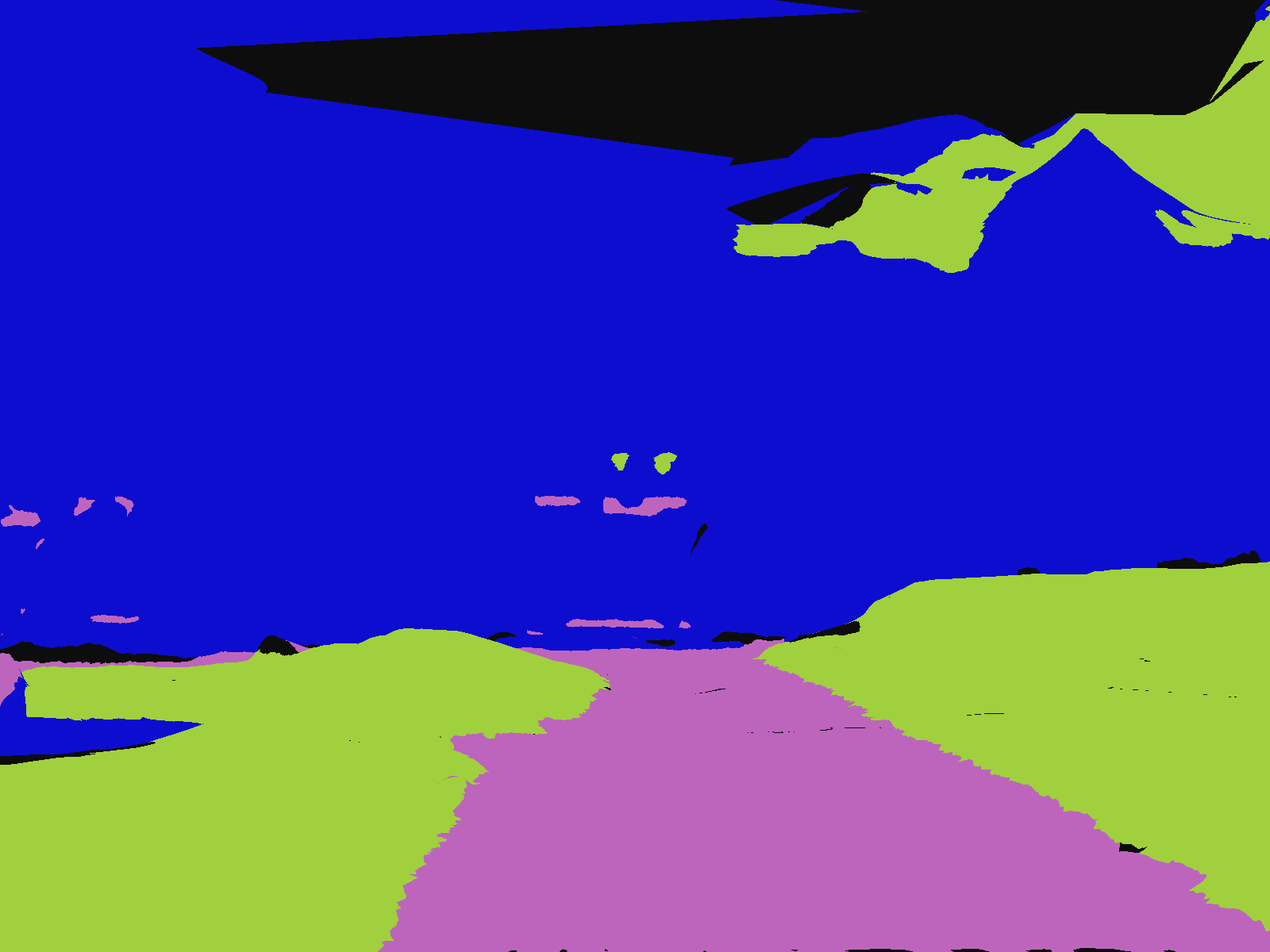}&
\includegraphics[width=0.18\textwidth,height=0.125\textwidth]{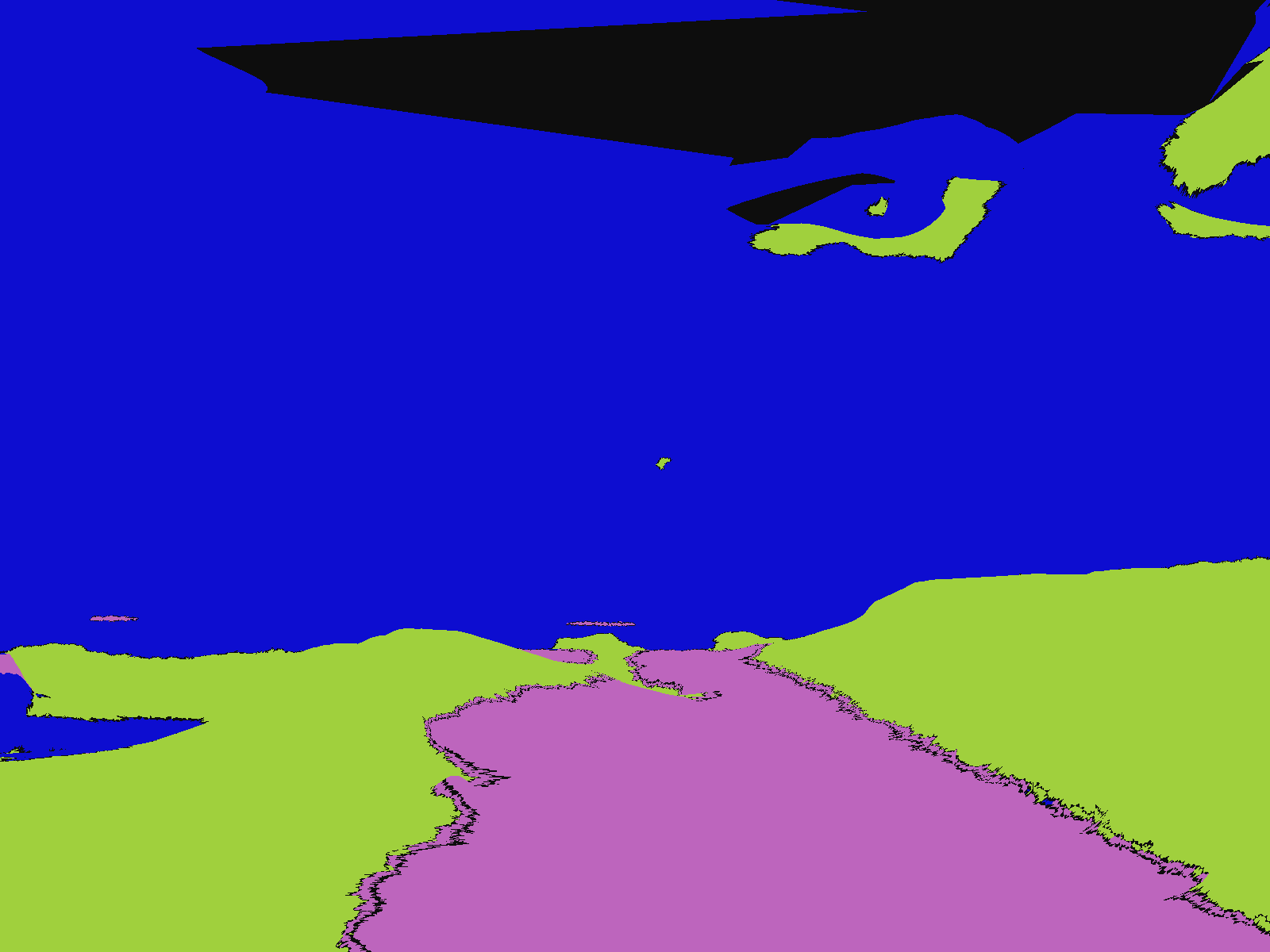}&
\includegraphics[width=0.18\textwidth,height=0.125\textwidth]{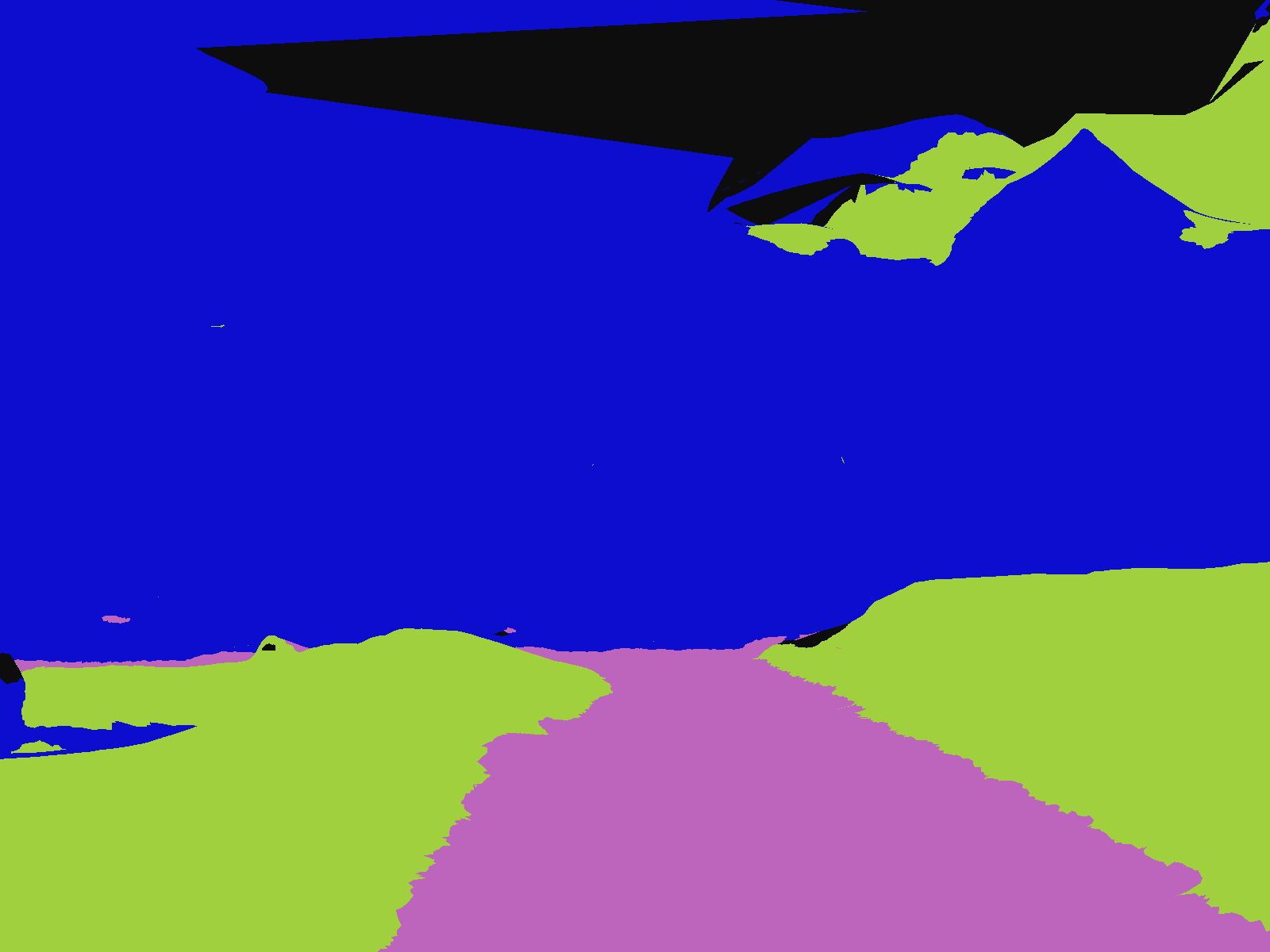}\\
\end{tabular}
\caption{Results on fountain-P11, castle-P30, Southbuilding, KITTI 95 and Dagstuhl}
\label{fig:results}
\end{figure*}

\begin{figure*}[ptb]
\centering
\setlength{\tabcolsep}{1px}
\begin{tabular}{cccc}
 Blaha \etal & Romanoni \etal &Proposed&\\
\cite{blaha2017semantically}&\cite{romanoni2017multi}  &&\\
\hline
\multicolumn{4}{c}{fountain-P11}\\
\includegraphics[width=0.25\textwidth]{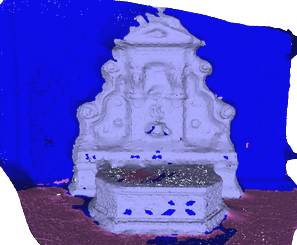}&
\includegraphics[width=0.25\textwidth]{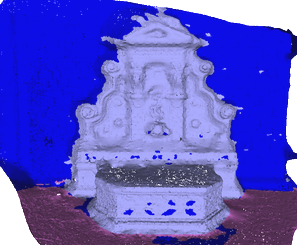}&
\includegraphics[width=0.25\textwidth]{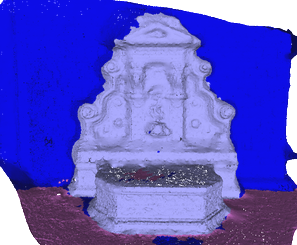}&
\includegraphics[width=0.08\textwidth]{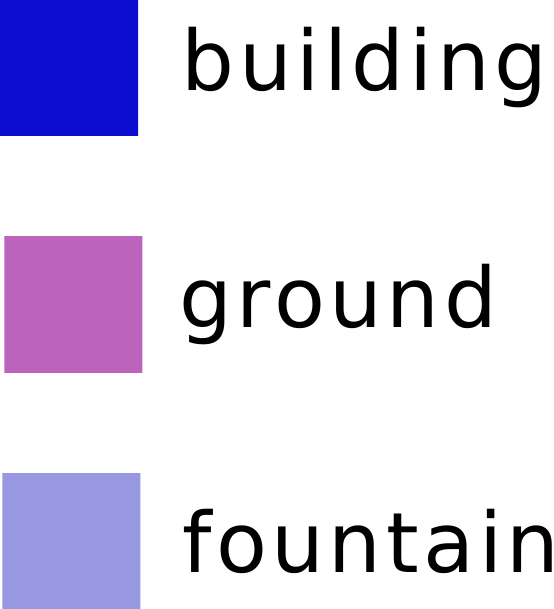}\\
\multicolumn{4}{c}{castle-P30}\\
\includegraphics[width=0.25\textwidth]{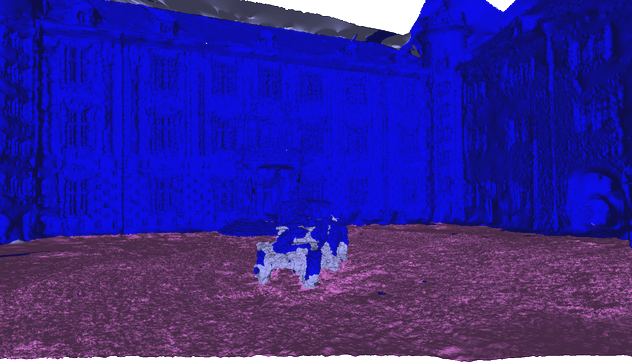}&
\includegraphics[width=0.25\textwidth]{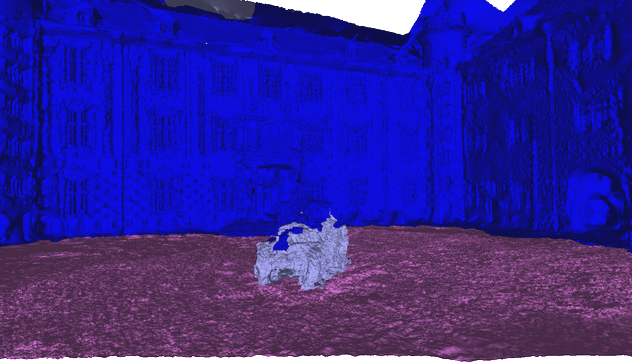}&
\includegraphics[width=0.25\textwidth]{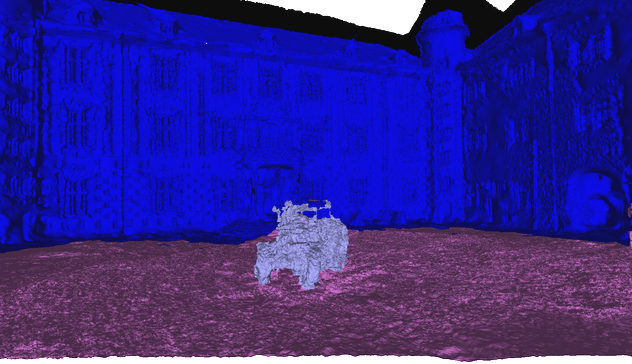}&
\includegraphics[width=0.08\textwidth]{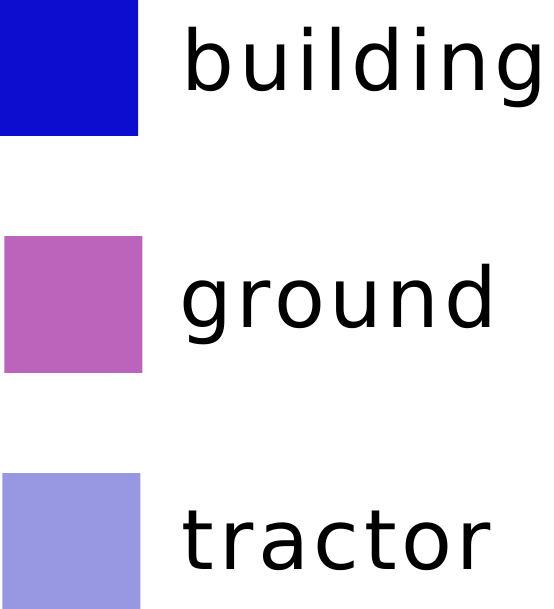}\\
\multicolumn{4}{c}{Southbuilding}\\
\includegraphics[width=0.25\textwidth]{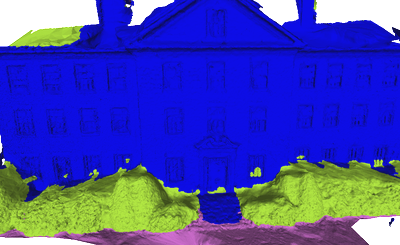}&
\includegraphics[width=0.25\textwidth]{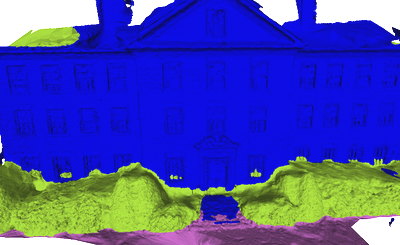}&
\includegraphics[width=0.25\textwidth]{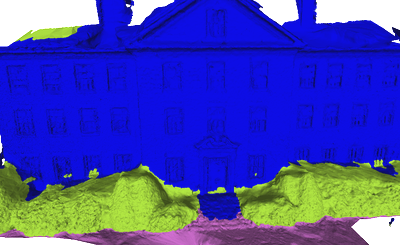}&
\includegraphics[width=0.1\textwidth]{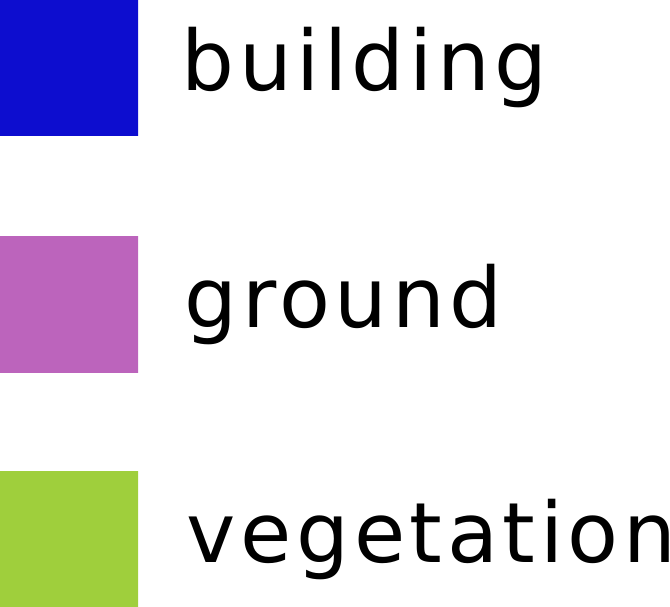}\\
\multicolumn{4}{c}{KITTI 95}\\
\includegraphics[width=0.25\textwidth]{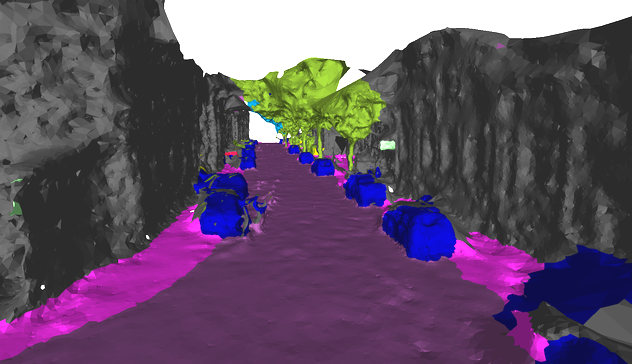}&
\includegraphics[width=0.25\textwidth]{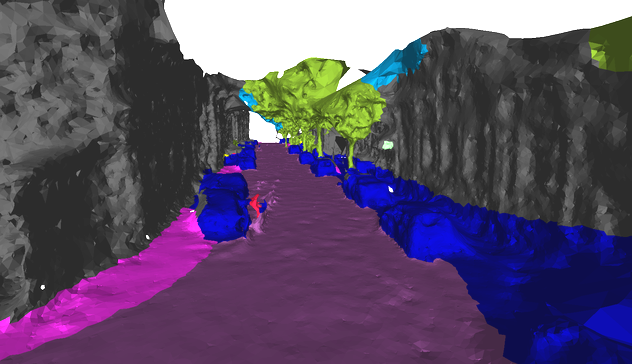}&
\includegraphics[width=0.25\textwidth]{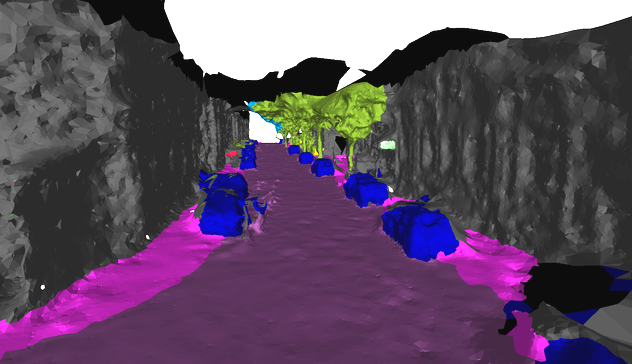}&
\includegraphics[width=0.1\textwidth]{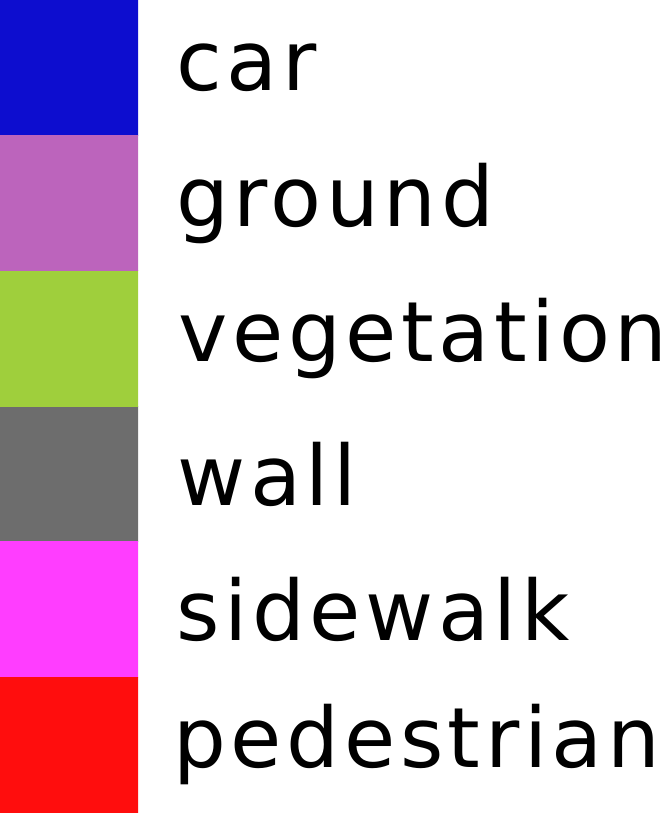}\\
\multicolumn{4}{c}{Dagstuhl}\\
\includegraphics[width=0.25\textwidth]{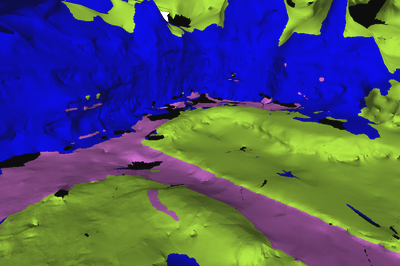}&
\includegraphics[width=0.25\textwidth]{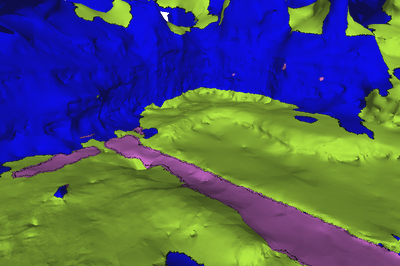}&
\includegraphics[width=0.25\textwidth]{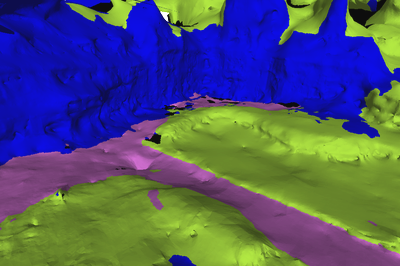}&
\includegraphics[width=0.08\textwidth]{./images/leg_south}\\
\end{tabular}
\caption{Labeled meshes}
\label{fig:resultsMesh}
\vspace{-8pt}
\end{figure*}

\begin{figure*}[tpb]
\centering
\setlength{\tabcolsep}{3px}
\begin{tabular}{ccc}
\includegraphics[width=0.28\textwidth]{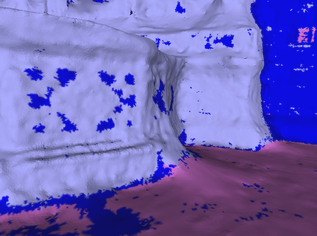}&
\includegraphics[width=0.28\textwidth]{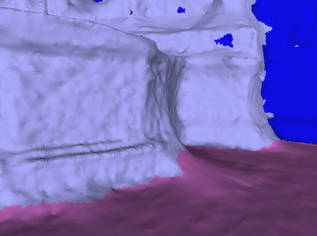}&
\includegraphics[width=0.28\textwidth]{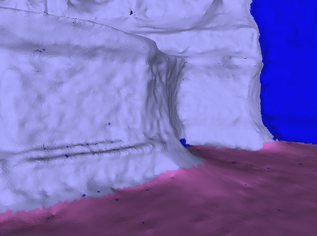}\\
Blaha \etal\cite{blaha2017semantically}& Romanoni \etal~\cite{romanoni2017multi}&Proposed\\
\end{tabular}
\caption{
Detail of the fountain-P11 dataset }
\label{fig:detail}
\vspace{-5pt}
\end{figure*}

\begin{table*}[tbb]
\centering
 \footnotesize
  \caption{Segmentation results in the failure case}
  \label{tab:resFail}
    \begin{tabular}{llccccccccc}
dataset & method& average & average & average & average & overall  &  overall  &  overall   & overall &  IoU\\
      & & accuracy  &  recall  &  F-score   & precision & accuracy  &  recall   &  F-score   & precision &  \\
\hline 
\multirow{4}{*}{DTU15}
&  [3]  &     {0.9518} & \textbf{0.7236} & 0.8579 & \textbf{0.7743} & \textbf{0.9569} & \textbf{0.8569} & 0.8645 & \textbf{0.8461} & \textbf{0.8462}  \\
& [17]  &     0.9517 & \textbf{0.7238} & {0.8261} & 0.7494 & 0.9552 & 0.8125 & 0.8715 & 0.8161 & \textbf{0.8460} \\
& Proposed  & \textbf{0.9547} & 0.6442 & \textbf{0.8948} & 0.6739 & 0.9444 & 0.7203 & \textbf{0.8989} & 0.7074 & 0.8552  \\
    \end{tabular}
\end{table*}

\begin{figure*}[tp]
\centering
\setlength{\tabcolsep}{1px}
\begin{tabular}{cccc}
\includegraphics[width=0.23\textwidth]{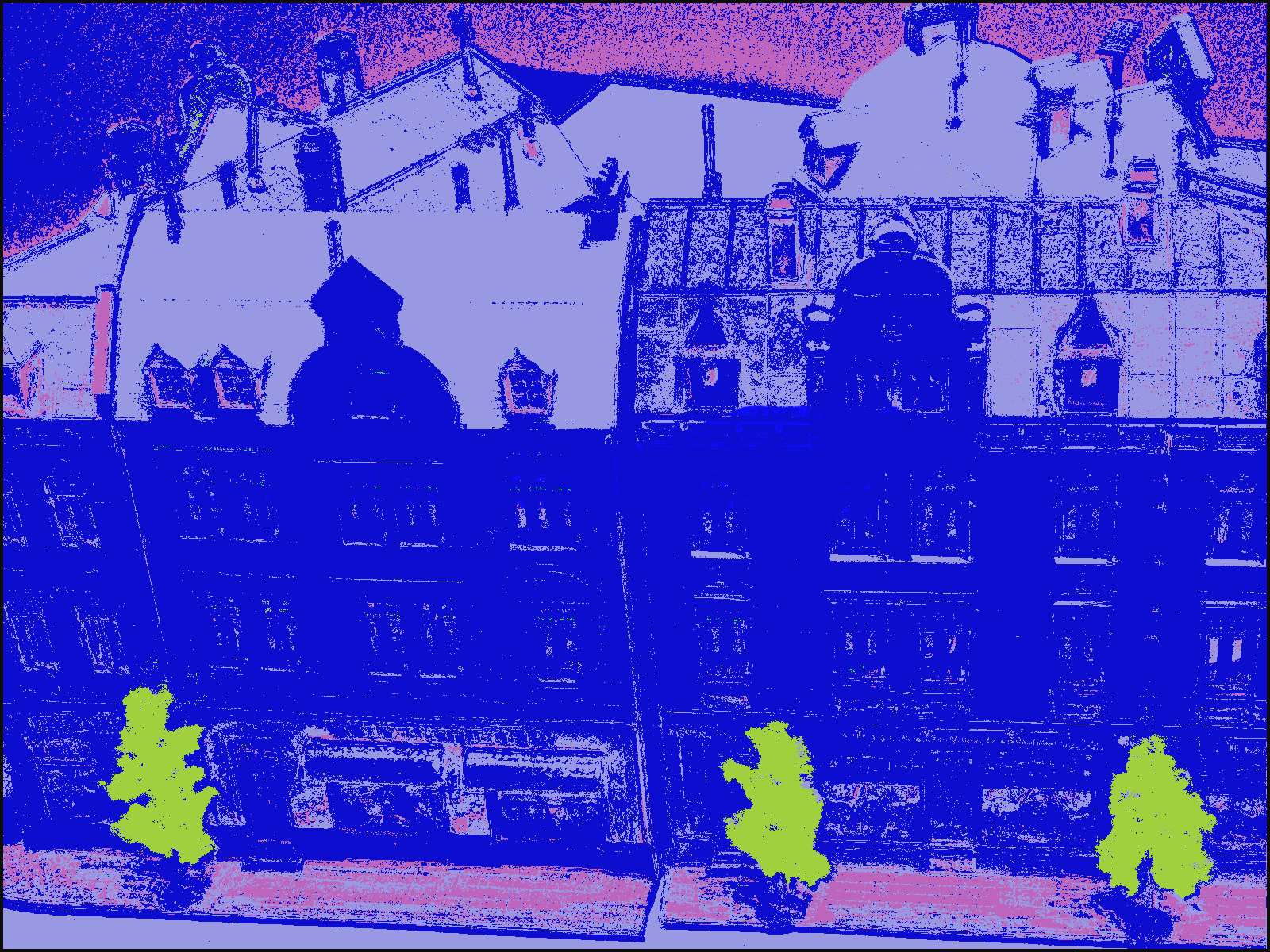}&
\includegraphics[width=0.23\textwidth]{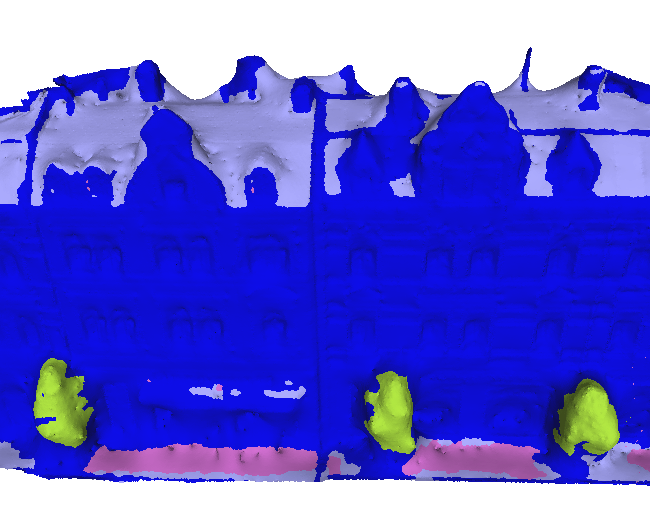}&
\includegraphics[width=0.23\textwidth]{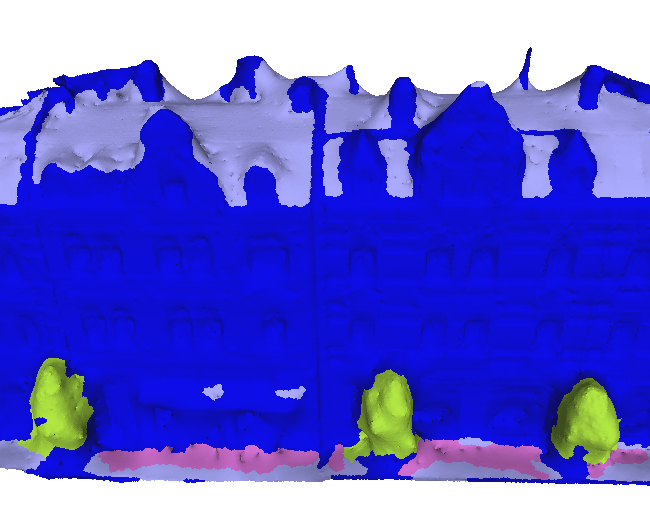}&
\includegraphics[width=0.23\textwidth]{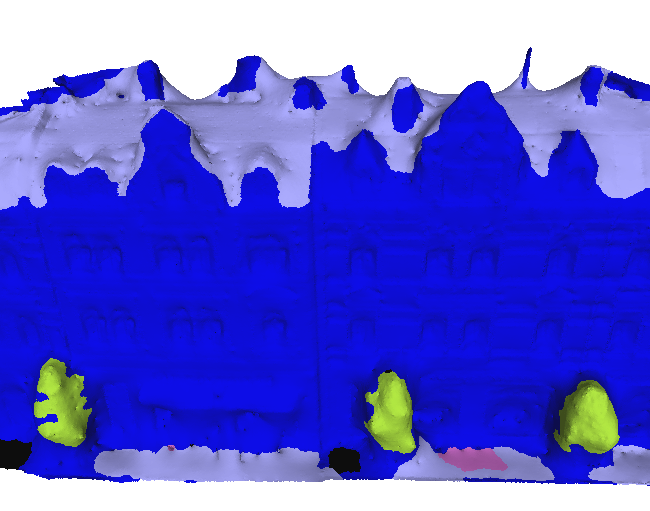}\\
Multiboost &Blaha \etal\cite{blaha2017semantically}& Romanoni \etal~\cite{romanoni2017multi}&Proposed\\
\end{tabular}
\caption{ \vspace{-5pt}
Failure case with the DTU15 dataset }
\label{fig:fail}
\vspace{-8pt}
\end{figure*}

We tested our method against the labeling algorithms in \cite{blaha2017semantically} and \cite{romanoni2017multi}.
Even if \cite{blaha2017semantically} and \cite{romanoni2017multi} iterate between mesh labeling and mesh refinement, and both steps indirectly influence each other, a single labeling step is a single block of the pipeline, and works independently from the refinement.
Since our method focuses on labeling, we compared it against the mesh labeling of \cite{blaha2017semantically} and \cite{romanoni2017multi} isolated from the refinement step.

As a first step, we tuned the three parameters $\mu_1$, $\mu_2$ and $\mu_3$ of the energy function \eqref{eq:fullene} on the dataset fountain-P11.
We assigned different values to each parameter while keeping the other two fixed.
We choose $\mu_1=0.2$, $\mu_2=0.2$ and $\mu_3=1.0$, and we use these same values with the other datasets.
To have a fair comparison we followed the same procedure to tune~\cite{blaha2017semantically} ($\mu_1=0.4$, $\mu_2=1.0$) and~\cite{romanoni2017multi} ($\mu_1=0.2$, $\mu_2=0.4$ and $\mu_3=0.4$).
For~\cite{blaha2017semantically} we also need to manually estimate the gravity vector.

While tuning the size of the cubes in the 3D lattice adopted to estimate the local distributions, we find that the parameter $d_i$ is quite robust to variations, it however depends on the dataset dimension and the labeling granularity and shows best results when it is not too big with respect to the scene, otherwise the results
tends to be similar to the global method proposed in~\cite{romanoni2017multi}. We choose $1.0\,$m, $7\,$m, $3.2\,$m, $4\,$m and $4\,$m respectively for the fountain-P11, castle-P30, Southbuilding, KITTI 95 and Dagstuhl datasets. In Figure \ref{fig:iouVoxel} we illustrate the intersection over union for the castle-P30 dataset with varying values of $d_i$ in comparison with the results for~\cite{blaha2017semantically} and~\cite{romanoni2017multi}.

In Table \ref{tab:resTot} we show the results of the proposed method against the methods in~\cite{blaha2017semantically} and~\cite{romanoni2017multi} and we evaluate the ablation of the proposed term $E_{\textrm{norm}}$ (Baseline).
We report different metrics: the average accuracy, recall, F-score and precision are computed as the average metric over all the pixel, while the overall measures are computed as the average over the classes.
To compute those metrics we compare the mesh rendering against 11, 15, 12, 5 and 4 different reference ground truth view respectively for the fountain-P11, the castle-P30, Southbuilding, KITTI 95 and Dagstuhl.
In Figure \ref{fig:results} we illustrate the segmentation results for a reference view  for the three datasets and in Figure \ref{fig:resultsMesh} the labeled mesh.
Even if the image segmentations coming from Multiboost are very noisy (first column of Figure \ref{fig:results}), the three algorithm managed to well compensate most of the misclassification.
Even if the method we propose does not require to add any prior information, its performance is better than the other methods in most of the metrics for each of the five datasets.
One of the reason is that, since the meshes have high resolution, the per-class distributions of the normals are quite complex, \eg, the ground is not simply flat, but it contains crispier details, therefore the simple prior of~\cite{blaha2017semantically} and the $P_{norm}$ term of~\cite{romanoni2017multi} are often not sufficiently expressive. Instead, our algorithm explicitly models these distributions through the histograms; therefore the prior-like normal term adapts itself to the scenario. Another reason of the improvement is evident in the fountain-P11 dataset. 
Here both~\cite{blaha2017semantically}, through the handcrafted prior, and~\cite{romanoni2017multi}, through its global prior, favor all the facets perpendicular with respect to the ground to be labeled as wall: this avoids to fill most of the 2D misclassification that happen on the fountain surface. Our approach completely overcome this issue since it collects the statistics locally, and not just because it acts as a strong smoothing term, indeed the small road sign in the KITTI 95 dataset is preserved (see Figure \ref{fig:resultsMesh}).
Finally, in addition to the processing shared with~\cite{blaha2017semantically} and~\cite{romanoni2017multi}, our method needs to compute the histogram; however the operation takes only 1.6 to 4.6 seconds.

In Table \ref{tab:resFail} we report a failure case that happened with the sequence 15 of the DTU dataset~\cite{aanaes2016large}. 
Here the Multiboost algorithm fails to provide a reliable segmentation of the ground (Figure \ref{fig:fail}). 
Therefore our method has not enough support to understand that facets directed upwards likely belongs to the ground.
The ground is better segmented  in~\cite{blaha2017semantically} because they enforce directly this concept by the class-specific prior and even in~\cite{romanoni2017multi} because they collect the normal statistics globally, therefore with a wider support.

\section{Conclusion and Future Works}
\label{sec:concl}
In this paper we proposed a novel method to label 3D models with semantic labels. 
In particular we defined a Markov Random Field over the mesh, as in the state-of-the-art, endowed with a new term that biases the  labeling depending on the facet orientation and acts as a class-specific geometric prior, but, instead of learning it or fixing  it manually, we estimate it directly from a coarse labeling of the mesh, collecting locally the distribution of the normals. 
Experiments on different datasets showed that the method is not only able to run without a learning phase or prior knowledge about the geometry of the classes, but it also improves the accuracy of the labeling.

As a future extension, we will investigate the possibility of running the method incrementally and on embedded multicores, coupling it with a real-time 3D reconstruction method~\cite{piazza2018real} also accounting the energy constraints imposed by the computing platform ~\cite{Zoni:2018:DEO:3212710.3186895,Zoni:2018:powertap}.
We also plan to test the method with a wider dataset such as ScanNet (\cite{dai2017scannet}), by replacing the Multiboost algorithm, that failed to estimate a reasonable segmentation of ScanNet scenes, with a more sophisticated image segmentation algorithm based on neural networks.
\section*{Acknowledgements}
{\small
This work has been supported by the ``Drones112'' project founded by EIT Digital. }

{\small
\bibliographystyle{ieee}
\bibliography{biblioTotal}
}

\end{document}